\documentclass[aps,pra,10pt,floatfix,superscriptaddress,twocolumn,nofootinbib]{revtex4-2}
\usepackage{graphicx}
\usepackage{amsmath,amsthm,amssymb,dsfont}
\usepackage{mdframed}
\usepackage{hyperref}
\usepackage{orcidlink}
\usepackage[capitalise]{cleveref}
\crefname{equation}{Eq.}{Eqs.}
\crefname{figure}{Fig.}{Figs.}
\usepackage{subcaption}
\usepackage{mathtools}
\usepackage{bm}
\usepackage{float}

\usepackage{silence}
\newcommand{\ket}[1]{\left| #1 \right\rangle}

\begin{document}

\title{Modelling Emotional Memory in Children with Tensor Networks}

\author{Henry Groves\,\orcidlink{0009-0004-7254-4063}}
\affiliation{Quantum Group, School of Computing, Newcastle University, 1 Science Square, Newcastle upon Tyne, NE4 5TG, UK}

\author{Lucia F. Jackson}
\affiliation{School of Psychology, Newcastle University, Wallace Street, Newcastle upon Tyne, NE2 4DR, UK}

\author{Barbara-Anne Robertson\,\orcidlink{0000-0002-6828-3022}}
\email{b.a.robertson@newcastle.ac.uk}
\affiliation{School of Psychology, Newcastle University, Wallace Street, Newcastle upon Tyne, NE2 4DR, UK}
\affiliation{Biosciences Institute, Newcastle University, Framlington Place, Newcastle upon Tyne, NE2 4HH, UK} 

\author{Jonte R. Hance\,\orcidlink{0000-0001-8587-7618}}
\email{jonte.hance@newcastle.ac.uk}
\affiliation{Quantum Group, School of Computing, Newcastle University, 1 Science Square, Newcastle upon Tyne, NE4 5TG, UK}

\begin{abstract}
We demonstrate how emotional valence influences the order-dependent structure of children's recognition memory: correct recall of a sequence of emotionally-valenced toys depended not just on the valence of a given toy itself, but also on the valence of the toys shown before and after it. Whilst standard psychological models confirm that order-dependence differs across an event (a set of toys shown in sequence), accuracy is low and the model does not reflect how memory for an emotional object influences others in the set. A classical tensor network model factoring in valence is able to achieve a 77.98\% accuracy in modelling the results of the study. While not strictly a ``quantum cognition'' model, this massive increase in accuracy shows the value of quantum-inspired methods for modelling order-dependent phenomena, such as emotional memory. Further, the task protocol we introduce presents a novel, real-world tool for exploring emotional temporal memory in children for analysis using classical and quantum-like models of cognition. 
\end{abstract}

\maketitle

\section{Introduction}

The brain does not store memories as a computer would, in fixed and objective snapshots of experiences. Memory recall is not simply a replay of a past event, but rather an active process where separate features (e.g., perceptual~\cite{MatherSutherland2011}, conceptual~\cite{Faul2025Vividness}, environmental/contextual~\cite{West2026Context}, and emotional \cite{West2026Context} that are bound together at encoding are reconstructed to comprise a representation of the past experience~\cite{schacter_adaptive_2012}. Memory is sensitive to the context in which encoding (creation) and retrieval (remembering) occurs. The process of rebuilding memories is not infallible, and can be subject to distortion and errors, whilst also being influenced by a person's current attitude, knowledge, and beliefs at the time of recollection~\cite{schacter_cognitive_2007}.

 Temporal memory operates as a specific facet of long-term associative memory, referring to the organisation of events within sequences~\cite{KesnerHunsaker2010} and there is a lack of consensus on how emotion impacts these processes~\cite{TalmiPalombo2025}. Childhood may be particularly informative in understanding how temporal order memory operates as functions such as the binding of elements and executive processes are still developing and remain highly sensitive to contextual influences~\cite{LoukesPrice2019}. Consistent with this, developmental vulnerability in temporal ordering accuracy has been linked to the maturation of key memory structures: the hippocampus, prefrontal cortex (PFC), large-scale connectivity networks~\cite{Bettencourtetal2021} and increasing functional connectivity and maturation of prefrontal and parietal systems~\cite{Rigginsetal2016} across childhood and adolescence. 

A foundational observation in human memory research is that emotional experiences are remembered more vividly and with greater accuracy than less emotional, more neutral experiences, a phenomenon referred to as the emotional memory enhancement effect~\cite{anderson_emotional_2006,bennion_oversimplification_2013,TalmiPalombo2025}. Early studies attributed the effect primarily to arousal (the intensity of the psychological activation of the experience), with the amygdala (a brain region critical to emotion) modulating structures that process memory during encoding and retrieval (i.e., the hippocampus)~\cite{mcgaugh_amygdala_2004}. More recent work has complicated this picture. Many analyses across recognition and free recall tasks have demonstrated that valence (the nature of the experienced emotion -- e.g., happy, sad, angry) \textit{and} arousal make separate contributions to memory storage and recollection~\cite{brainerd_semantics_2019}. Ref.~\cite{brainerd_semantics_2019} also show that the influence of valence scales quadratically, whereas arousal affects memory linearly. This appears to be due to different pathways being used in the brain for different stimuli. Neural imaging using fMRI scanning has identified different neural pathways for arousal-driven and valence-driven enhancement: an amygdala and hippocampus network for the former, and a prefrontal cortex and hippocampus network for the latter~\cite{ritchey_level_2011}. In general, much of the literature relies on neutral stimuli, does not clearly define when emotional valence is positive, negative, or a combination of the two, and assumes linear encoding, potentially limiting its ability to capture context-sensitive and order-dependent retrieval. 

 According to the arousal-biased competition theory, emotionally salient stimuli dominate attentional processing, strengthening item representations~\cite{MatherSutherland2011}. However, this enhancement may come at a cost: emotion can narrow attentional scope, improving item memory while impairing the encoding of contextual relationships such as temporal order. Whilst robust item-memory effects are demonstrated, order-memory outcomes are inconsistent when arousal and valence are not clearly dissociated~\cite{schmidt_emotions_2011}. Further, models that treat recall as binary (remembered/forgotten) are implicitly assuming a categorisation which may overly simplify complex processes. The consensus in the literature is that the greater the arousal, the more attentional resources are narrowed, following Ref.~\cite{easterbrook_effect_1959}'s cue utilisation hypothesis. However, there is no solid answer as to what constitutes `central' vs `peripheral' information~\cite{lanciano_memory_2011, payne_impact_2006}, and whether arousal alone is the cause of this narrowing, or if discrete emotional categories also matter~\cite{kaplan_motivation_2012}. These inconsistencies highlight a limitation of classical models in accounting for context-sensitive and order-dependent dynamics of sequential recall.  

 Few studies have examined how emotional valence and repetition jointly shape children’s temporal order memory within a formal framework capable of modelling order-dependent interference~\cite{TalmiPalombo2025,Deker2021PolarBear}. Existing work prioritises item recall over sequencing, and relies on classical linear models that assume fixed encoding and stable retrieval, limiting their ability to explain variability and order effects in children’s recall. Children often struggle to accurately sequence neutral information, yet show improved recall when events carry emotional salience~\cite{ArterberryAlbright2020}. However, classical accounts do not adequately explain why emotionally-enhanced encoding may coexist with inconsistent or order-sensitive recall patterns, nor how repetition interacts with valence in shaping temporal reconstruction.

Whilst it is known that emotion impacts temporal order memory, it is poorly understood how these effects arise~\cite{petrucci_matter_2021}. Studies of temporal order under strong emotional conditions have reported both impairments~\cite{wang_emotional_2025} and enhancements~\cite{dev_negative_2022,schmidt_emotions_2011}, dependent on experimental conditions. If the direction (improving or worsening recollection of the temporal order) of the effect depends on conditions such as arousal intensity, emotional valence or the sequence in which experiences occur in, this implies that temporal order memory is not just another simple component of recall. Instead, it suggests that the temporal ordering of memories are intertwined with the emotional context of experiences. Therefore, a model aiming to reflect how human memory functions cannot simply treat memories as a sequence of independent events. It must be able to capture the relationships between order and emotional context to be a successful model.

This dependency between order and emotionality is also shown when looking at the primacy and recency effects. These are well-established effects where people recall the first items (primacy) and last items (recency) in a sequence better as opposed to the items in the middle of the sequence~\cite{Glanzer1966Recall}. These effects are a good starting point for studying temporal structure in recall, but importantly, these effects do not appear to be stable when combined with emotional context~\cite{giovannelli_emotional_2022}. The presentation of emotional images disrupted expected recall effects for those at the start and end positions, whereas emotional images presented during the middle of the sequence reduced how many were forgotten~\cite{giovannelli_emotional_2022}. This suggests that the emotional valence of experiences competes with, and reshapes positional encoding, rather than just improving or worsening a person's ability to recall. It therefore appears necessary to model position and emotional content jointly and not independently to create a realistic model. Emotional valence does not simply scale recall probabilities uniformly, it instead appears to restructure the dependencies across an event.

A more complete model of emotional memory should therefore:
\begin{enumerate}
	\item Represent the distribution of probabilities across positions, not just individual probabilities per item.
 	\item Be sensitive to the emotional context of items and the order in which they appear.
 	\item Be capable of representing context-dependent structure between recall events.
 \end{enumerate}
 
 Classical probabilistic models, which typically assume some degree of independence between items, are based on commutative mathematics, which is structurally ill-suited to meet these requirements. A model that treats each recall independently cannot reproduce this conditional structure, regardless of how many parameters it has.
 
 The Quantum Formalism, and mathematical tools built upon it, such as Quantum Instruments and Tensor Networks, appear to be more naturally suited to represent the kind of structured, order-dependent and contextually-correlated behaviour we see in emotional memory data. It therefore offers an alternative framework for modelling context-sensitive and order-dependent processes. Importantly, this does \emph{not} posit quantum physical processes in the brain; rather, it applies the Hilbert space structure used to obtain probabilities in quantum mechanics to cognition, offering a coherent account where judgements are contextual, states evolve through prior evaluations, and interference arises between representations~\cite{Wangetal2013}. Unlike classical models, quantum approaches permit cognitive states to exist in probabilistic superpositions, where multiple potential representations coexist until a judgement occurs, collapsing the state and altering subsequent probabilities \cite{Heisenberg1958,Pothos_Busemeyer_2013}. Retrieval can therefore be conceptualised as a state-changing process rather than extraction of a fixed memory trace. 

 These ideas are supported by order effects that challenge classical assumptions. For example, the Question Order Effect (QOE) involves responses to an initial question influencing responses to a subsequent one~\cite{Rasinskietal2012}, and the Response Replicability Effect (RRE) involves repeated judgements altering response probabilities. These effects demonstrate dependence on prior evaluations, violating independence assumptions~\cite{OzawaKhrennikov2022}. Quantum models account for these patterns through principles including noncommutativity ($AB\neq BA$), where evaluation order affects outcomes, and interference, whereby evaluating one element shifts the probability structure of another~\cite{Tsuchiyaetal2025}. Empirical examples, including the Clinton-Gore effect in political research, illustrate how sequential evaluation shapes responses~\cite{fisher2021act}.

 Extending this framework, quantum models have also been applied to interactions between affective and rational processes. Affective and cognitive components can be represented as interacting vectors within a probabilistic state, allowing interference between evaluative and rational components~\cite{Whiteetal2016}. Rather than contributing additively, affective evaluation can modify judgement probabilities through interference, producing effects not predicted by classical models. From this perspective, children’s temporal memory may reflect the collapse of a probabilistic configuration shaped by contextual and emotional influences rather than retrieval of a fixed sequence. This highlights the need to examine how valence and repetition jointly influence order-dependent recall.

 While quantum models for emotional memory potentially meet these requirements, they fall foul of their own issues. The quantum cognition models given previously~\cite{busemeyer_quantum_2012,White2014Ask,Whiteetal2016,ozawa_modeling_2021,OzawaKhrennikov2022, fuyama_quantum-like_2025,busemeyer_incorporating_2025} represent human memory as vectors in a Hilbert space, and measurements (such as asking a question of a subject) as operators acting on that space. For a single item or a short sequence of binary questions, this is computationally tractable. However, as the number of items grows, a fundamental problem emerges that motivates the use of tensor networks.

 If we want a quantum model to represent the joint probability distribution of a sequence of $n$ items, with each item having $d$ outcomes, then the total number of parameters required to describe this system is $d^n$. This is necessary to represent every single possible combination of outcomes. With just 2 outcomes and 5 items, the number of combination is 32. With 3 outcomes and 5 items, the number of combinations is 243. By 6 outcomes, the number climbs to 7776. With longer experiments or more complex recall tasks with more outcomes, which are more analogous to human experience, this number grows exponentially.

 This problem is referred to as the \textit{curse of dimensionality}, and it is an obstacle for any model trying to explicitly represent a complete joint distribution. Classical models can avoid this by assuming independence of items, as, if each item's recall is not reliant on any others, then only $n$ parameters are needed. But, as discussed above, human memory recall is not a system of independent events, and making this assumption discards exactly the correlations we are trying to capture when modelling memory. A more principled approach is needed, one that can represent complex, structured correlations without requiring an exponential number of parameters. One potential solution for this is classical tensor networks~\cite{white_density_1992,vidal_efficient_2003}, specifically, Matrix Product States (MPS)~\cite{perez-garcia_matrix_2007, orus_practical_2013}, which enable representations of order-dependence without requiring a full $d^n$-parameter model.

 In this paper, we develop such a model, based on tensor networks, and apply it to model data from a novel study looking at recall of emotionally-valenced images in children. We show that this model is able to achieve a $77.98\%$ accuracy in modelling the results of our study, showing the efficacy of such an order- (and valence-) dependent model over standard psychological models of memory.

 This paper is laid out as follows. In~\Cref{sec:Experiment}, we describe a study we performed using a task we developed to obtain raw data on children's ability to recall emotionally-valenced information. Specifically, this study involved showing 50 children a sequence of 5 toys, each of which were either valenced neutral (wooden blocks of different shapes and sizes) or valenced positive (toy cars, action figures, etc). Not only was it recorded whether the child correctly recalled a given toy as being part of the sequence, but also whether the child got the toy's position in the sequence correct.

In~\Cref{sec:Class}, we present a standard psychological analysis of the experimental results, using two-way within-subjects' repeated-measures ANOVAs, and $t$-tests. As expected, these standard psychological tools fail to adequately demonstrate the inherent order-dependence of the event, instead just showing overall mean differences with high levels of noise. We also give a classical model, which as expected fails to adequately model the event, giving high inaccuracy.

In~\Cref{sec:basictn}, we discuss a simple approach to modelling the experiment using tensor networks. We introduce the tensor network formalism, then describe how the study is encoded as a probability tensor. In this preliminary model, each item in a trial is represented by a binary variable (either recalled or not recalled). We then detail the matricisation procedure and sequential singular value decomposition used to factorise this tensor into an MPS, and explain how the bond dimensions emerging from this process can be used as a direct measure of inter-item dependencies in recall.

In~\Cref{sec:tnprediction}, we introduce a method to use the model discussed in~\Cref{sec:basictn} for prediction. For a given item, we condition the MPS by projecting known observations onto the previous tensors and marginalise unknown future sites, reducing the network to a local scoring problem at the target item. The predicted recall outcome is then taken as the most probable state under this conditioned distribution.

In~\Cref{sec:qutrittn}, we extend the binary model to use a qutrit representation, expanding each item's label space from two to three dimensions, to distinguish between totally incorrect recall, correctly remembering the item but in the wrong order, and remembering both the correct item and order. This richer encoding better reflects the experimental data. The MPS construction and prediction procedure remain largely unchanged, only accounting for the larger state space.

In~\Cref{sec:valence}, we introduce the final model of this paper, which directly incorporates the emotional valence of each item into the tensor network. Each site's physical index is expanded to encode both valence and recall outcomes as a single composite state, forcing the model to learn their joint distribution, rather than treating them as separable. During prediction, the known valence of the target item conditions the target tensor onto the relevant recall subspace, to more accurately reflect how emotional context operates in the actual experiment.

Finally, in \Cref{sec:Discussion}, we summarise our findings, discuss limitations with our study and model, and then present paths for future works to build upon the results we present.

\section{Experimental Protocol}\label{sec:Experiment}

\textbf{Participants:} Participants (N = 50) were children  aged four to eleven years ($M_{age}$ = 6.9, SD = 2.2 years; 22 females, 20 males, 8 unspecified) recruited from Newcastle University’s \textit{Science Adventures} family-based science outreach events running in the School of Psychology during October and February half-term holidays, with data combined into a single dataset. Participation was voluntary, producing a self-selecting sample. Eligibility required participants to fall within the specified age range at testing. Parents reported developmental or learning profiles, which were noted but not assessed. 

Ethical approval was granted by Newcastle University’s Faculty of Medical Sciences Ethics Committee (ethics code 2457\_6). Both the primary researcher and research assistant held enhanced DBS checks. Informed parental consent was obtained prior to participation, and children provided verbal assent. Participants were informed that they could withdraw at any time without consequence. Participants exercised this right; 24\% withdrew after Trial 5 (TR1), primarily due to fatigue or boredom, while 62\% completed all ten trials. Negative stimuli were not included to ensure that even temporary negative emotional states were not evoked in the children. 

\textbf{Design:} The study employed a $2\times2$ within-subjects design examining emotional valence and repetition effects on children’s temporal order memory. All participants completed all levels of the independent variables. Emotional valence was manipulated within-subjects', with participants experiencing both positive and neutral objects in each stimulus set, enabling comparison. Eight stimulus sets comprised five neutral and positive toys (see Figure 1 which shows Set 4 with three neural and two positive objects). Stimulus set presentation was randomised over trials such that each child had a unique set order assigned to trials. Repetition was operationalised by re-presenting two object sets later in the task, allowing re-encoding before recall. Repeat trials re-presented the original sequence. Each Trial was coded using specific labels (see \Cref{tab:TrialEncoding}). 

\begin{table}
    \centering
    \begin{tabular}{c|c}
\hline
Trial Code & Description\\
\hline
T1 & Encoding Trial 1\\
T2 & Encoding Trial 2\\
T3 & Encoding Trial 3\\
T4 & Encoding Trial 4 \\
TR1 & Retrieval Trial 1 (repeat of T1)\\
T5 & Encoding Trial 5\\
T6 & Encoding Trial 6\\
T7 & Encoding Trial 7\\
T8 & Encoding Trial 8\\
TR2 & Retrieval Trial 2 (repeat of T5)
    \end{tabular}
    \caption{Trial Encoding Scheme used in the Study. Note. T refers to encoding Trials assessing QOE, and TR refers to Retrieval Trials assessing RRE. Stimuli were organised into sets of 5 objects, sets were randomly presented across trials such that each child had a unique set order across trials.}
\label{tab:TrialEncoding}
\end{table}

Dependent variables captured episodic and temporal recall performance. These included object selection accuracy, order accuracy (number of items in the correct serial position), recall errors, and decoy selection. Ordering response time (s) was recorded using a
stopwatch.

Following~\cite{Bettencourtetal2021}, two decoy objects were included in each set to ensure that children intentionally selected objects they had seen. Selecting only target objects indicated intentional encoding, whereas selecting a decoy indicated an error. Only two children selected a decoy, each in a single trial, these responses were excluded from analysis.

\textbf{Materials:} 60 children's toys were used as experimental stimuli. Age-appropriate items were selected to be similar in size but unique enough to be discriminated based on shape, colour, and perceptual complexity. See examples in Figure 1. Of these, 20 objects were categorised as emotionally neutral, consisting of plain wooden and neutral-coloured blocks. 20 objects were categorised as positive and visually engaging. The remaining 20 toys served as decoys. Target objects were grouped into eight stimulus sets, each containing a mixture of positive and neutral items. To control for order and learning effects, stimuli set presentation was randomised across participants using a random number generator, ensuring each participant received a unique sequence, minimising familiarity bias. Additionally, to assess recall consistency and RRE, the sequence from Trial 1 was repeated after Trial 4 (TR1), and the sequence from Trial 5 was repeated after Trial 8 (TR2) (see Table 1). 

\begin{figure}[t]
    \centering
\includegraphics[width=1.0\linewidth]{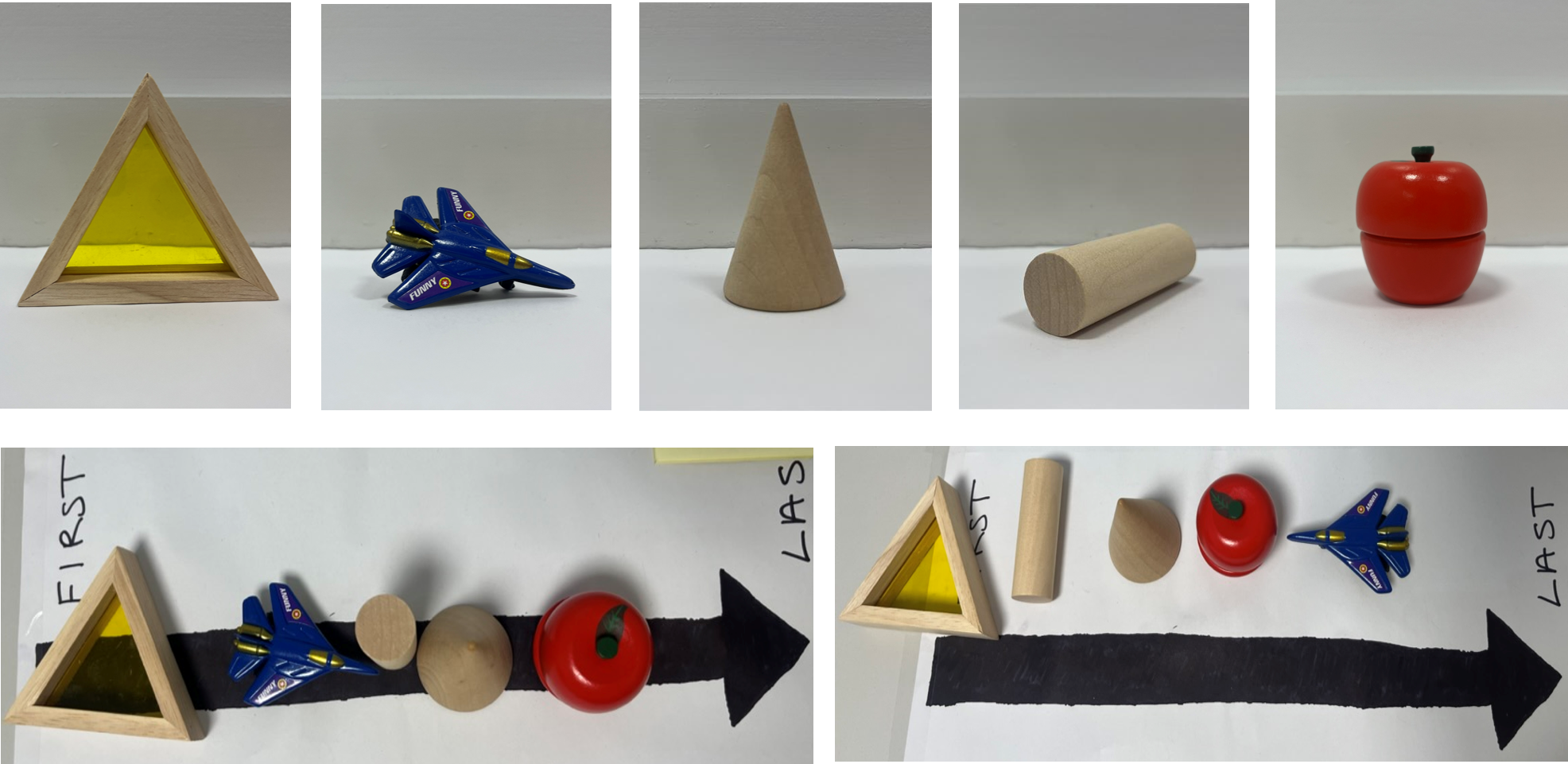}
    \caption{The top row displays objects used in stimulus set 4 in the order they were presented at encoding. The bottom row shows two retrieval phases from two different children. Note that both children selected the correct items but made different errors in temporal ordering when asked to recall the objects in the exact order they had previously seen. Children were given the black arrow showing where the first and last items should be placed, this procedure was established during the practice trial to ensure all children understood how to complete the task. Not shown: two decoy toys that were included in the toy box to ensure intentional selection.}
    \label{fig:placeholder}
\end{figure}

Operationalising emotional valence in children presents methodological challenges because stimulus preferences vary across individuals. There is no database of validated, emotionally-valenced children's toys, so a supporting aim of this project was to determine if these stimuli were emotionally engaging to children. Negatively valenced stimuli were not used due to ethical considerations~\cite{Bettencourtetal2021}. The design therefore assessed whether the intended emotional manipulation was effective by examining whether positively valenced objects produced measurable recall differences relative to neutral stimuli. 
 
Pictures of the stimuli were photographed at a distance of 30 cm away under standardised conditions (uniform white background, consistent angle; see Figure 1 for examples) and uploaded to PsychoPy~\cite{Peirce2019}, where a custom script-controlled stimulus presentation, sequencing, and timing was created for each stimuli set. Although stimulus timing was electronically controlled, manual recording of completion times was necessary due to the physical manipulation component of the task. This combination of computerised stimulus control and manual measurement ensured both experimental precision and ecological validity.

\textbf{Procedure}: Testing was conducted individually in a quiet cubicle in the School of Psychology. Participants sat approximately 47 cm from a computer screen and received standardised, age-appropriate verbal instructions to view object sequences and later, select and arrange them in the order seen at presentation.

Participants completed a practice trial to ensure task comprehension. During the practice trial, three objects were presented sequentially on PsychoPy~\cite{Peirce2019}. Participants then retrieved the corresponding objects and one decoy object not present in the presentation set in a toy box. They were asked to arrange the objects they had seen along a directional arrow in presentation order. The performance criterion was complete accuracy in the task. Practice trials were repeated until this criterion was met. The children were allowed to repeat the practice trial to reach the performance criterion three times before being excluded. Three children repeated the practice task once before achieving full accuracy. This ensured that all participants understood task requirements.

The study consisted of ten trials. In each trial, five objects were presented individually for 10 seconds each, with a 5 second inter-stimulus interval. Shorter stimulus presentations have been used (2000ms with a 500ms fixation cross)~\cite{Bettencourtetal2021}; however, developmental research indicates children require longer encoding durations. A mask (centrally presented black fixation cross on a grey screen) was shown between each presentation to interrupt residual sensory traces and ensure that each stimulus was processed independently~\cite{Sligteetal2008}. Upon completion of the task, children were praised and given a stamp in their Science Adventures ‘passport’.

Following the presentation of each set, participants were presented with a toy box containing seven objects: five target objects and two decoys. Participants were instructed to select the objects they had just seen and arrange them along a directional arrow from the first to the most recently presented object. Completion times were recorded manually using a stopwatch. The final arrangement was photographed for offline coding. Photographs ensured accurate scoring of recall, order accuracy, and decoy selection. Data were subsequently coded from photographs and entered into Microsoft Excel for analysis. See Figure 1. 

\section{Classical Analysis}\label{sec:Class}

\subsection{Classical Data Analysis Methodology} 

Temporal Order memory was operationalised as the ability to reconstruct the sequence of five target objects presented within each trial. Order accuracy was scored based on the serial position in which each object was placed along the directional arrow, with one point awarded for each object in the correct position (maximum score = 5). Object recall accuracy was scored as the number of correctly selected target objects regardless of placement. Scores were recorded in Microsoft Excel and compared with the original presentation sequence to calculate the number of correctly recalled positive and neutral objects, both in correct order and irrespective of order. All scoring was performed manually. Data were analysed using IBM SPSS Statistics (version 31.0.1.0 (49)). All tests were two-tailed with alpha set at 0.05.

To examine QOE, Repeated-Measures (RM) ANOVAs were conducted for object recall and order memory. Within-subjects factors were Trial (T1-T8) and Valence (positive vs. neutral). Analyses assessed main effects and their interaction. When sphericity was violated, the Greenhouse-Geisser correction was applied (G-G corrected). Polynomial contrasts were conducted to examine linear trends across trials. Estimated marginal means for Trial, Valence, and their interaction were compared using Bonferroni-adjusted comparisons.

To examine RRE, planned pairwise comparisons using Paired-Samples t-tests compared performance between the initial and repeated trials. Trial 1 was repeated after approximately five minutes, with three intervening trials (TR1). Similarly Trial 5 was repeated as the tenth presentation (TR2). Separate tests compared trials for positive and neutral conditions. Whilst 62\% of children completed all 10 trials, 96\% of participants completed the task to TR1. To test if the increased power between T1 - TR1 altered the findings, a supplementary RM ANOVA examined Trials 1-4 and TR1. These data are not shown as the results did not differ from those in complete dataset. These results are available, along with data on completion time on request.

An exploratory analysis examined whether recall accuracy varied as a function of prior emotional context. For each item, the number of previously presented positively-valenced items was calculated, and accuracy at each serial position was analysed as a function of this value to assess the influence preceding emotional content on recall.

Due to experimenter error, data for Trials 2 and 3 were missing for Participant 19. Although the child completed the trials, photographs were not taken for offline scoring. Missing values were replaced with the participant’s mean score (positive $M = 1.25$; neutral $M = 0.75$), justified by the child completing all ten trials. The same procedure was applied for Participant 22 in Trial 1 (positive: $M = 2.1$, neutral: $M = 2.1$).

\subsection{Results of Classical Analysis}

\textbf{Positively-valenced objects are remembered better than neutral objects}: A RM ANOVA examined the effects of the two within-subjects' variables: Trial (T1-T8) and Valence (positive vs. neutral) on object recall scores. The main effect of Trial was not statistically significant, $F (3.90, 113.15) = 2.12$, $p = 0.085$. There was a significant main effect of valence, $F (1, 29.00) = 4.79$, $p = 0.037$, indicating that objects of positive valence ($M = 2.44$, $SD = 0.09$) were better recalled than neutral objects ($M = 2.37$,
$SD = 0.10$). The Trial x Valence interaction was
not significant, $F (3.45, 203.00) = 0.50$, $p = 0.707$ (G-G corrected), indicating that valence-related differences were consistent over trials but linear contrasts did revealed a significant decrease in performance across trials, $F (1, 29) = 7.17$, $p = 0.012$.

\textbf{Temporal order performance is superior for positively valenced objects}: A RM ANOVA was conducted to examine the effects of Trial (T1-T8), representing first-presentation order, and Valence (positive vs. neutral) on order memory scores ($N = 30$). There was no significant main effect of Trial, $F (7, 203.00) = 1.72$, $p = 0.105$, indicating no difference in performance over time. There was a significant main effect of Valence, $F (1, 29.00) = 7.44$, $p = 0.011$, positively valenced objects were ordered more accurately than neutral objects. The interaction was not significant, $F (4.79, 138.75) = 0.34$, $p = 0.879$ (G-G corrected), indicating that order performance across trials did not differ by valence. 

\textbf{Response replicability effect not visible:}   Paired-samples’ $t$-tests compared original trials with their corresponding repeated trials to examine RRE on temporal order accuracy. T1 was compared with its repeat (TR1), and T5 with its repeat (TR2). Analyses were conducted separately positive and neutral valence conditions.

For the comparison between T1 and TR1, no significant differences were observed in either condition. For T1, temporal order accuracy did not significantly differ between the original and repeated presentations in the positive condition, $t (47) = 0.86$, $p = 0.393$, or the neutral condition, $t (47) = -0.46$, $p = 0.651$. A similar pattern was observed for Trial 5, with no significant differences between original and repeated trials in the positive condition, $t (30) = -0.68$, $p = 0.502$, or the neutral condition, $t (30) = -0.19$, $p = 0.851$. Overall, temporal order performance remained stable across repetitions for both valence types.

\section{Tensor Networks}\label{sec:basictn}

Tensors are the building blocks of tensor networks. A tensor is a multidimensional array of numbers. The number of indices to specify a value in a tensor is equal to the rank of the tensor.

Contraction is the operation that connects the nodes of a tensor network together. Contracting over two indices is how information is passed between tensors, and can be seen as communication channels, or in our model, as dependencies (``entanglement'') between specific events being modelled by such tensors.

A Matrix Product State (MPS) is a type of tensor network which represents a high-order tensor as a chain of lower-order tensors, connected by shared indices. In the high-order tensor, all $d^n$ entries must be stored explicitly. An MPS factorises it into $n$ rank-3 tensors~\footnote{Boundary nodes are rank-2 in an open MPS, though for simpler computation they are treated as having a third trivial rank.}, $A^{(1)}, A^{(2)}, \ldots, A^{(n)}$, each of which is constructed from correlations in the data rather than every possible outcome. For a rank-$n$ tensor $T_{i_1,i_2,\dots,i_n}$, the MPS factorisation takes the form:
\begin{equation}
T_{i_1,i_2,\dots,i_n}=\sum_{\mathclap{\alpha_1,\dots,\alpha_{n-1}}}\;A_{i_1,\alpha_1}^{(1)}A_{\alpha_1,i_2,\alpha_2}^{(2)}\cdots A_{\alpha_{n-1},i_n}^{(n)}\,.
\end{equation}

Indices $\alpha_1,\dots,\alpha_{n-1}$ are the virtual indices, and connect adjacent tensors when contracted. Indices $i_1,\dots,i_n$ are the physical indices, and correspond to the real-life outcomes we're modelling. The size of each virtual index is referred to as the bond dimension $\chi$, and is the variable that determines the expressivity of an MPS. It determines how much information is passed between adjacent tensors, and in the proposed model of human memory, how strongly the past recall outcomes can influence the future ones. A bond dimension of 1 means no information is passed between sites at all, which would be equivalent to the assumption in classical models that recall outcomes are independent. A bond dimension of $\chi > 1$ allows correlations to propagate through the network, with larger $\chi$ permitting richer and more complex dependencies. This will provide a metric to determine if the dependencies and interactions in human memory can be captured using this methodology, as the bond dimensions describes exactly that.

\begin{figure}[t]
    \begin{center}
		\includegraphics[width=\linewidth]{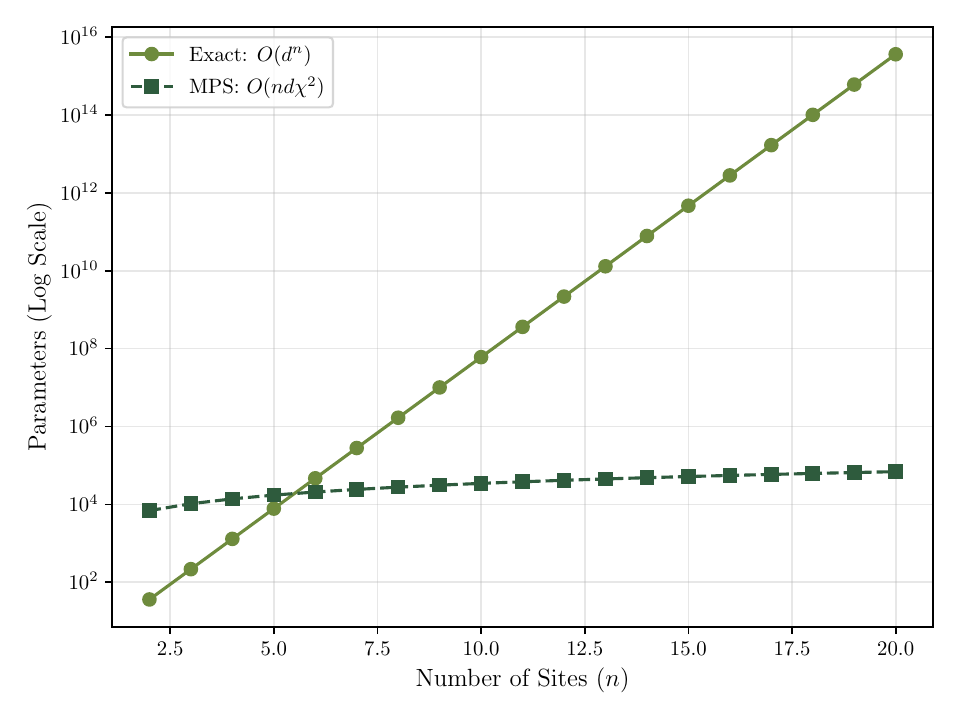}
	\end{center}
	\caption{Scaling of parameters required to model the same system with an exact representation and an MPS representation.}\label{fig:mpsvsexact}
\end{figure}

The total number of parameters in the MPS scales as $\mathcal O (n\cdot d\cdot\chi^2)$ rather than $\mathcal{O}(d^n)$, which is its key advantage as the size of systems we are trying to model, and their number of outcomes, grows (see~\Cref{fig:mpsvsexact}).

\subsection{Modelling a trial}

For convenience and consistency with the tensor network literature, we use a Dirac-style basis notation, writing the recall outcomes at position $i$ in terms of the basis states $\ket{0}_i$ and $\ket{1}_i$:
\begin{align*}
	\ket{0}_i & \to\text{incorrect recall at position }i\,, \\
	\ket{1}_i & \to\text{correct recall at position } i\,,
\end{align*}
such that each items outcome is indexed by a two dimensional label space $\mathcal{H}_i\cong\mathbb{C}^2$, and a trial of $n$ items is indexed by the composite label $\mathcal{H}_\text{total}=\bigotimes_{i=1}^n\mathcal{H}_i$.

Following standard tensor network convention, we expand a trial's recall outcomes over this composite basis as
\begin{equation}
    \ket\Psi = \sum_{\mathclap{i_1,\dots,i_5}}\;T_{i_1 i_2 i_3 i_4 i_5}\ket{i_1}\otimes\ket{i_2}\otimes\cdots\otimes\ket{i_5},
\end{equation}
where $\ket\Psi$ is not a quantum state, but a representation of the full joint distribution over recall outcomes: The coefficients $T_{i_1\dots i_5}$ are real, non-negative probabilities rather than complex amplitudes.

By treating the experimental results as a multidimensional array (a tensor), we can capture the `probability landscape' of this specific memory task, which can be used to determine the dependencies (represented as bond-dimension structure, and implying non-Markovianity rather than physical entanglement~\cite{Guo2020BondasNonMarkoianity}) that simple classical models fail to predict. Successfully mapping these dependencies is a necessary step in demonstrating that quantum-like frameworks are a valid approach for modelling order effects in human memory. Each index in $T_{i_1i_2i_3i_4i_5}$ corresponds to one of the $2^n$ recall outcomes. $T$ represents the global probability landscape of the trials, where:
\begin{equation}
\begin{split}
T&_{i_1,i_2,i_3,i_4,i_5} =\\
&P(X_1 = i_1,\ X_2 = i_2,\ X_3 = i_3,
X_4 = i_4,\ X_5 = i_5).
\end{split}
\end{equation}
The tensor is populated from the experimental data as follows:
\begin{equation}T_{i_1, i_2, i_3, i_4, i_5} = \frac{1}{N} \sum_{k=1}^{N}\prod_{m=1}^5\delta_{s_m^{(k)} i_m}
,\label{eq:dataload}\end{equation}
where $N$ is the total number of trials in the data set, $s^{(k)}$ represents the sequence observed in the $k$-th trial, $i_m \in \{0, 1\}$ for $m = 1,\dots,5$,
and the following normalisation constraint holds:
\begin{equation}
\sum_{i_1,\dots,i_5}T_{i_1,\dots,i_5}=1.
\end{equation}

\subsection{Matricisation}\label{sec:3matricisation}

As discussed previously, the bond dimension $\chi$ in tensor networks is the measure of expressivity of the tensor network. Specifically, it quantifies the amount of correlation between parts of the quantum system. A bond dimension of 1 implies no correlation between nodes and the network reduces to a simple product state. $\chi$ being greater than one represents the presence of correlations between the objects, indicating that the order of recall, and the items before, affects a person's ability to recall when modelling human memory using a tensor network approach. We use this value to determine if a quantum approach is valid to model the order effects in human memory.

Singular Value Decomposition is used to determine the bond dimension for between nodes. For a given matrix $\mathbf{M}$, SVD is a factorisation of the form:
\begin{equation}
	\mathbf{M}=\mathbf{U}\bm\Sigma \mathbf{V^\dagger}\,,\label{eq:3.8svd}
\end{equation}
where $\mathbf{U}$ is a unitary matrix representing a rotation, $\bm\Sigma$ is a diagonal matrix with non-negative real numbers on the diagonal, and $\mathbf{V^\dagger}$ is the adjoint of another unitary matrix representing another rotation. 
This enables us to do two things: First, we can isolate pathways of correlation. In the context of this model, the columns of $\mathbf U$ represents the patterns of the `past' memory states, and the rows of $\mathbf{V}^\dagger$ represent the possible `future' outcomes. $\bm\Sigma$ acts as a bridge between the two, with its values quantifying the weight of each corresponding rank-1 component from $\mathbf{U}$ and $\mathbf{V}^\dagger$.  Secondly, these values can show us which rank-1 components actually represent structural patterns, and which are negligibly small and most likely just represent statistical noise. Therefore, we can compress the amount of information required without incurring too much error, and to calculate the bond dimension. 

To perform SVD on the tensor $T$ that represents the outcome of the trial, we must first reconfigure the tensor into a 2D matrix, as SVD is a linear algebra technique and requires a matrix with rows of observations and columns of features. Bipartite Matricisation is applied to the tensor to achieve this. 
If we consider the recall of item $k$, we can split the remaining items into two blocks representing the past and the future. The past is represented by items $1..k$, and the future by $k+1..n$, where $n$ is the total number of items in a given trial (5 in our case).

\Cref{fig:3.1matricised} shows the fully populated matrix $\mathbf{M}^{(2)}$ after a split at $k=2$, with example probabilities. We can see that the tensor has been transformed into a `look-up table' between past and future. Given a certain past recall, we can then see the probabilities that future recall patterns would occur.
\begin{table}
\centering
		\begin{tabular}{cc|cccc}
			Index  &                           & 1         & 2         & $\dots$  & 8         \\
			       & Past/Future & $\ket{000}$ & $\ket{001}$ & $\dots$  & $\ket{111}$ \\
			\hline
			1      & $\ket{00}$                  & 0.0003    & 0.004     & $\dots$  & 0.15      \\
			2      & $\ket{01}$                  & 0.041     & 0.001     & $\dots$  & 0.46      \\
			$\vdots$ & $\vdots$                    & $\vdots$    & $\vdots$    & $\ddots$ & $\vdots$    \\
			4      & $\ket{11}$                  & 0.005     & 0.0356    & $\dots$  & 0.26
		\end{tabular}
	\caption{The matricised tensor $\mathbf{M}^{(2)}$ after bipartite matricisation at $k=2$}\label{fig:3.1matricised}
\end{table}

\subsection{Singular Value Decomposition}\label{sec:svd}
Once the tensor is matricised at a given split, we apply SVD following \Cref{eq:3.8svd}. The matrix $\mathbf{U}$ represents the basis of the past, defining all the memory states the subject can be in after the items they have already recalled. $\mathbf{V^\dagger}$ represents the basis of the future, defining how each of the past states map onto the potential recall outcomes of the remaining items.
The diagonal values in $\bm\Sigma$ are referred to as the singular values, and are sorted in descending order, where:
\begin{equation}
\sigma_1\ge\sigma_2\ge\dots\ge\sigma_r >0\,,
\end{equation}
with $r$ being the rank of the matrix $\mathbf{M}$. These singular values represent the amount of correlation between past and future states and results. Writing as a sum of outer products:
\begin{equation}	\mathbf{M}=\sum_{i=1}^r\sigma_i\mathbf{u}_i\mathbf{v_i^\dagger},\quad r=\operatorname{rank}(\mathbf{M})\,,
\end{equation}
$\mathbf{u}_i$ represents the `past' singular vectors which capture patterns that the memory system has experienced up to the point of the past/future split. $\mathbf{v_i}$ represents the `future' singular vectors, which encode how specific memories and memory states inform future behaviour and outcomes.

The outer product $\mathbf{u}_i\mathbf{v_i^\dagger}$ creates a rank-1 matrix that isolates a specific mapping between past and future. The corresponding singular value $\sigma_i$ quantifies how strong the effect between a given past recall pattern and future pattern is. The matrix $\mathbf{M}$ is then formed from the superposition of these weighted rank-1 components.

\subsection{Determining the bond dimension}\label{sec:bondd}

The bond dimension $\chi$ is derived directly from the singular value matrix $\bm\Sigma$. 
For example, if $\sigma_1$ is large and the other singular values are near zero, we take the bond dimension as 1, implying there is no correlation between nodes.

When modelling experimental data, many of the lower-magnitude singular values are typically negligible and represent empirical noise rather than true structural patterns. Therefore, to calculate the effective $\chi$ for a matricised tensor $\mathbf{M}$, we must establish the number of rank-1 layers required to represent the underlying matrix without overfitting to noise.
The Eckart-Young-Mirsky~\cite{eckart_approximation_1936,mirsky_symmetric_1960} theorem provides the theoretical foundation for this truncation. It states that the optimal approximation $\tilde{\mathbf{A}}$ of a matrix $\mathbf{A}$ is guaranteed by retaining the $\chi$ largest singular values, such that the Frobenius error is minimised:
\begin{equation}
\tilde{\mathbf{A}}=\sum_{i=1}^\chi \sigma_i \mathbf{u}_i\mathbf{v}_i^\dagger =\operatorname*{argmin}_{\text{rank}(\mathbf{A}^\prime)\leq \chi} \big|\big|\mathbf{A}-\mathbf{A^\prime} \big|\big|_F\,.
\end{equation}
In practice, the bond dimension is determined by establishing a cutoff threshold $\epsilon$, and retaining only the singular values that exceed it. Hence, the bond dimension $\chi$ is given by
\begin{equation}
    \chi=\sum_i\delta_{\sigma_i>\epsilon}\,,
\end{equation}
where 
\begin{equation}
    \delta_{\sigma_i>\epsilon}=\begin{cases}1 \text{ if } \sigma_i > \epsilon \\ 0 \text{ if } \sigma_i \leq \epsilon\end{cases}
\end{equation}

\subsection{Sequential Singular Value Decomposition}\label{sec:ssvd}

To determine the bond dimensions between all nodes in the tensor network, this process of matricisation and SVD is repeated for each of the four possible cuts in the network, turning the tensor $T$ into a MPS. This allows us to find the bond dimensions in one deterministic pass, with no optimisation loop required like other methods, e.g., density matrix renormalisation group algorithm (DMRG)/tensor cross interpolation (TT-cross).\footnote{Note however that this does produce a locally rather than globally optimal result, meaning there may be other MPS representations of the same tensor which use fewer parameters. Sequential Singular Value Decomposition (SSVD) is computationally simple, but the bond dimensions it produces is an upper bound, rather than necessarily optimal.}

The method is as follows:

\textbf{Cut 1}
Begin by separating the first item $i_1$ from the rest of the sequence.
\begin{enumerate}
	\item Reshape $T$ into a matrix $\mathbf{M}^{(1)}$ of size $2\times 16$:
	      $$\mathbf{M}^{(1)}_{(i_1),(i_2 i_3 i_4 i_5)}=T_{i_1,i_2,i_3,i_4,i_5}$$
	\item Perform SVD on $\mathbf{M}^{(1)}$:
	      $$\mathbf{M}^{(1)}=\mathbf{U}^{(1)}\bm\Sigma^{(1)}\mathbf{V}^{(1)\dagger}$$
	\item Determine $\chi_1$:
	      $$\chi_1=\sum_i\delta_{\sigma_i>\epsilon}$$
	\item Define the first node of the MPS, $A^{(1)}$, as the past singular vectors, where $\alpha_i$ is the virtual index connecting the next node of dimension $\chi_i$:
	      $$A^{(1)}_{i_1,\alpha_1}=\mathbf{U}^{(1)}_{i_1,\alpha_1}$$
	\item Calculate the residual matrix, $R^{(1)}$, by multiplying the singular values into the future singular vectors. This contains the dependent (``entangled'') future of the sequence.
	      $$R^{(1)}_{\alpha_1,(i_2 i_3 i_4 i_5)}=(\bm\Sigma^{(1)}\mathbf{V^{(1)\dagger}})_{\alpha_1, (i_2 i_3 i_4 i_5)}$$
	      The size of $R^{(1)}$ is $\chi_1\times16$.
\end{enumerate}

\textbf{Cuts 2--4}
The process is repeated iteratively. At each cut $k$, the next physical index $i_k$ is absorbed into the past block by reshaping the residual $R^{(k-1)}$ into a matrix $\mathbf{M}^{(k)}$ of size $(\chi_{k-1}\cdot 2)\times 2^{5-k}$, performing SVD, and extracting the MPS node $A^{(k)}$ from the left singular vectors. At the final cut, no further decomposition is possible, so the remaining residual directly yields the last node $A^{(5)}$.

We now factorise our 5th order probability tensor $T$ into a sequential contraction of five local tensors, also known as an MPS:
\begin{equation}
\begin{split}
	T&_{i_1, i_2,i_3,i_4, i_5}=\\
    &\sum_{\{\alpha\}}A_{i_1,\alpha_1}^{(1)}A_{\alpha_1, i_2,\alpha_2}^{(2)} A_{\alpha_2, i_3,\alpha_3}^{(3)} A_{\alpha_3, i_4,\alpha_4}^{(4)}A_{\alpha_4,i_5}^{(5)}\,.\label{eq:3.13mps}
\end{split}
\end{equation}
The tensors $A^{(1)}$ and $A^{(5)}$ are defined with boundary indices $\alpha_0$ and $\alpha_5$ respectively, and due to being open boundaries (simply the tensors on either end of the chain), the bond dimensions are set to $\chi_0=\chi_5=1$.

We can visually represent \Cref{eq:3.13mps} using Penrose's graphical notation~\cite{penrose_applications_1971}. This is a visual language to represent the flow through tensor contractions. The indices of each tensor are represented by connecting lines (referred to as `legs'). Connected legs in the diagram represent contraction.

\begin{figure*}
	\begin{center}
		\includegraphics[width=0.75\linewidth]{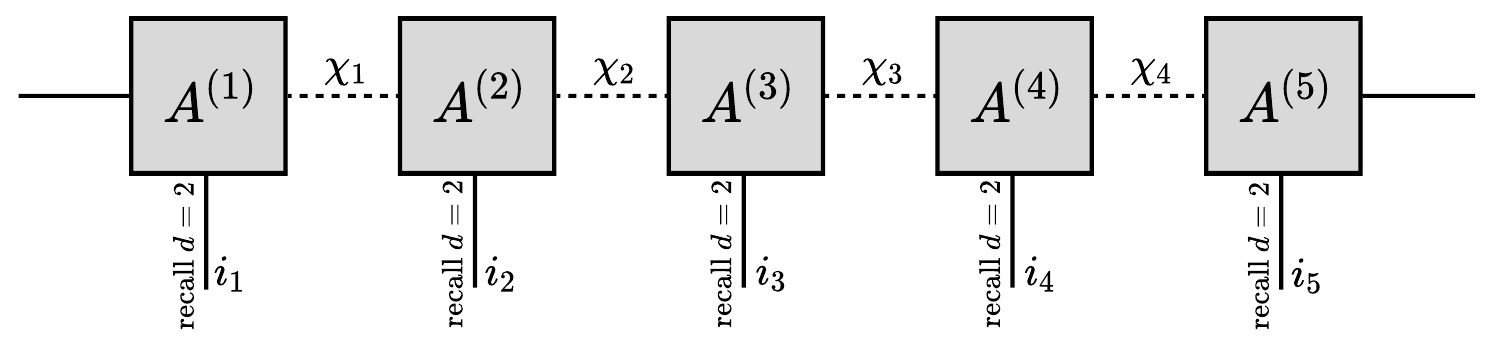}
	\end{center}
	\caption{A diagrammatic representation of the MPS from \Cref{eq:3.13mps} in Penrose graphical notation.}\label{fig:mps_as_notation}
\end{figure*}

\subsection{Applying to Experimental Data}

To load our experimental data into this model, we follow~\Cref{eq:dataload}. In the implementation, Laplace smoothing is applied to all entries in the tensor through a uniform prior $\lambda=0.01$:
\begin{equation}
T^{0}_{i_1,i_2,i_3,i_4,i_5}=\lambda,\forall i_n\in\big\{0,1\big\}.
\end{equation}
This ensures that if a specific recall sequence was not seen in the study, it can still be predicted. If no smoothing was applied, and there were 0-probabilities in the probability tensor (which is almost guaranteed due to the sparsity of data), multiplying by a zero-state would collapse the entire tensor network to zero, making prediction impossible. The tensor is then filled with values
\begin{equation}
T_{i_1,i_2,i_3,i_4,i_5}=\lambda+\frac{1}{N} \sum_{k=1}^{N}\prod_{m=1}^5\delta_{s_m^{(k)} i_m}\,,
\end{equation}
and scaled by its Frobenius norm ($\mathcal{L}_2$ Norm) to stabilise the values for SVD compression
\begin{equation}\tilde{T}=\frac{T}{||T||_F}\,.
\end{equation}
This normalisation is necessary as SVD algorithms are sensitive to the scale of values inside the matrix, and scaling $T$ down to a unit hypersphere prevents potential exponential `blow-up' when multiplying tensors together.

We then follow the sequential singular value decomposition process (\cref{sec:ssvd}) to factorise $\tilde{T}$ into the form shown in \Cref{eq:3.13mps}. The maximum bond dimension is $\chi=4$ and singular values are discarded using an $\epsilon$ value of $1\times10^-4$.\footnote{In \Cref{sec:TNDataAnalysis} we show the effect of sweeping this $\epsilon$ discard value from $10^{-5}$ to $10^-1$ on each model's prediction accuracy.}

\section{Prediction using a Tensor Network}\label{sec:tnprediction}

While simply representing the experimental data with this model is useful for demonstrating the presence of dependencies on past states (``entanglement'') in human memory, we can extend these tensor network approaches beyond observation to predict future recall states.

To predict the outcome of recall of a specific target item $t\in\{1,\dots,5\}$ from a given trial, we condition the model at the node $t$ and then evaluate the marginal distribution.

\subsection{Projection of Known Observations}

To condition on the past observations, we project the known observations into the network. We `pin' the physical indices to the previous observations, $k\in\{1,\dots,t-1\}$, by taking the inner product of each tensor representing the past items, $A^{(k)}$, with the standard basis vector, $\mathbf{e}^{(i_k)}$, that represents the observed state of that item. As the physical output dimension is $d=2$, $\mathbf{e}^{(i_k)}$ collapses to one of the following basis vectors:
\begin{align*}
\mathbf{e}_{\text{incorrect}}=\ket{0}=\begin{pmatrix}1&0\end{pmatrix}^T,\\\mathbf{e}_\text{correct}=\ket{1}=\begin{pmatrix}0&1\end{pmatrix}^T.\end{align*}
Each tensor is reduced from a rank-3 tensor (representing the left virtual bond, the physical index, and the right virtual bond), to a rank-2 tensor (simply a matrix) $B^{(k)}$:
\begin{equation}\label{eq:projection}
	B^{(k)}_{\alpha_{k-1},\alpha_k}=\sum_{\mathclap{s_k\in\{0,1\}}}\;A^{(k)}_{\alpha_{k-1}\,, s_k,\alpha_{k}}\cdot\mathbf{e}^{(i_k)}_{s_k}.
\end{equation}

The rank-3 tensor $A^{(k)}$ can be thought as a stack of matrices, where each matrix corresponds to a specific physical state and is referred to as a `slice'. In the case of this model, there are two physical states $s_k\in{0,1}$, and therefore we have two matrices in the stack. To more cleanly represent this process of slicing, we use `slice notation', where 
\begin{equation}
    B^{(k)}_{\alpha_{k-1},\alpha_k}=A^{(k)}_{\alpha_{k-1},i_k,\alpha_k}=\left[A^{(k)}(i_k)\right]_{\alpha_{k-1}, \alpha_k}.
\end{equation}
Here, $i_k$ represents the actual observed state index. $B^{(k)}$ is now a transfer matrix, a fixed operator that takes a vector of information from the previous operators, transforms it, and passes it to future operators.

Note that while $A^{(1)}$ (and similarly $A^{(5)}$) is depicted as a rank-2 matrix in~\Cref{eq:3.13mps}, here it is treated as a rank-3 tensor $A^{(1)}_{\alpha_0,i_1,\alpha_1}$, where the virtual indices $\alpha_0$ and $\alpha_5$ are restricted to 1. This is to simplify computation as all the sites in the MPS can be processed the same way.

\subsection{Marginalising the Future}
Marginalisation is the process of removing the influence of variables in our tensor network. As the model is not yet aware of the outcome of the nodes in the future, we must calculate every future possibility, their associated probabilities, and then sum them together, which erases the future's influence on the item we're currently trying to predict. To trace out these future sites, we contract their physical indices with a vector of ones,
$$\mathbf{v}=\begin{pmatrix}1&1\end{pmatrix}^T,$$
which again collapses the rank-3 tensors to rank-2 transfer matrices:
\begin{align}B^{(k)}_{\alpha_{k-1},\alpha_k}&=\sum_{\mathclap{s_k\in\{0,1\}}}\;A^{(k)}_{\alpha_{k-1}, s_k,\alpha_{k}}\cdot\mathbf{v}_{s_k}\,.
\label{eq:conditioning}\end{align}

\subsection{Local Scoring and Prediction}
To predict the recall outcome of our target item $t$, we must determine the probability that each outcome occurs, and select the most likely of these.

We construct the left environment, $E_L$, which represents the past observations, through the sequential product over the shared indices $\{\alpha_1,\dots\alpha_{t-2}\}$ of all the transfer matrices we have constructed:
\begin{equation}
	[E_L]_{\alpha_0,\alpha_{t-1}}=\sum_{\alpha_1,\dots,\alpha_{t-2}}\prod_{k=1}^{t-1}B^{(k)}_{\alpha_{k-1},\alpha_k}\label{eq:leftenv}\,.
\end{equation}
All shared indices of $E_L$ (the virtual bonds of the tensors that make up the past) have been contracted, leaving only the boundary index ($\alpha_0$), and the index shared with the site we're predicting ($\alpha_{t-1}$) open.

We then construct the right environment, $E_R$, which represents the potential future recall outcomes, again by sequential product over the shared indices $\{\alpha_{t+1},\dots,\alpha_{n-1}\}$:
\begin{equation}
	[E_R]_{\alpha_t,\alpha_n}=\sum_{\alpha_{t+1},\dots,\alpha_{n-1}}\prod_{k=t+1}^{n}B^{(k)}_{\alpha_{k-1},\alpha_k}\label{eq:rightenv}\,.
\end{equation}
Again, looking at the indices of $E_R$, the virtual bonds of the tensors that make up the future have been contracted, leaving only the index shared with the site we're predicting ($\alpha_t$) and the boundary index ($\alpha_n$) open.

The term $A^{(t)}_{\alpha_{t-1},s_t,\alpha_t}$ is our target tensor, whose physical index $s_t$ is left open to be evaluated. We calculate a score for each potential outcome of $s_t$, summing over every remaining index. This produces a rank-1 tensor $R$ that holds the score for each recall outcome:
\begin{equation}
\begin{split}
	R&_{s_t}=\\
    &\sum_{\alpha_0,\alpha_{t-1},\alpha_{t},\alpha_{n}}[E_L]_{\alpha_0,\alpha_{t-1}} A^{(t)}_{\alpha_{t-1},s_t,\alpha_t} [E_R]_{\alpha_t, \alpha_n}\label{eq:scoring}\,.
\end{split}
\end{equation}
With the scores computed for each recall outcome (correct/incorrect), the prediction is made by choosing the state that is most likely:
\begin{equation}
	\hat{s_t}=\operatorname*{arg max}_{s_t\in\{0,1\}}|R_{s_t}|\,.\label{eq:argmaxselection}
\end{equation}
Whilst the index form in \Cref{eq:leftenv,eq:rightenv,eq:scoring} is useful in showing how the indices contract and how we sum over all indices leaving just $s_t$, we instead can re-write \Cref{eq:leftenv,eq:rightenv} in pure matrix form, abstracting away the $\alpha$ indices:
\begin{equation}
	E_L = \prod_{\mathclap{k=1}}^{t-1}\;B^{(k)}, \qquad E_R = \prod_{\mathclap{k=t+1}}^{n}\;B^{(k)},
\end{equation}
and we can rewrite \Cref{eq:scoring} as:
\begin{equation}
	R(s_t)=E_L A^{(t)}(s_t)E_R\,,
	\label{eq:matscoring}
\end{equation}
where $A^{(t)}(s_t)$ represents the two-dimensional matrix for a given state of $s_t\in\{0,1\}$.


\section{Qutrit Tensor Network}\label{sec:qutrittn}

Whilst this initial method of modelling emotional memory shows that there are dependencies (``entanglement'') between nodes, it is deliberately oversimplified. The first step in improving this model would be changing how the experimental results are classified. In the previous model, even if a subject successfully recalled an item they had seen but just in the incorrect order, that would be classed as $\ket{0}$ and treated as incorrect.

\subsection{Modelling a trial}\label{sec:trialqutrit}

Instead, let us represent the recall as a three-dimensional label space, $\mathcal{H}\cong\mathbb{C}^3$ as opposed to a two-dimensional space, with the basis states:
\begin{align*}
	\ket{0}_i & \to\text{Incorrect item and position for item }i,      \\
	\ket{1}_i & \to\text{Correct item, incorrect position for item }i, \\
	\ket{2}_i & \to\text{Correct item and position for item }i.
\end{align*}

As the dimension of the label space has increased, the size of our 5th-order tensor $T$ is now $3\cdot3\cdot3\cdot3\cdot3$ for a total of $3^5=243$ possible recall-position patterns. The tensor is populated using \Cref{eq:dataload}, resulting in:
$$T_{i_1,i_2,i_3,i_4,i_5}, i_n\in\big\{0,1,2\big\}.$$

\subsection{Matricisation}\label{sec:4matricisation}
Again, to perform SVD, we must matricise this tensor, exactly the same way as previous. The only change is the base of the super-indices, to 3.

\subsection{Qutrit Matrix Product State}
The MPS for this qutrit model is constructed in a similar way as outlined in \Cref{sec:ssvd}, using Sequential Singular Value Decomposition. The same four cuts are made with the same residual passing logic, the only difference being in the size of the matrices throughout, as the physical dimension of each contracted tensor is now $d=3$ rather than $d=2$. We write this as a similar factorisation to as \Cref{eq:3.13mps}, with the only difference being each physical index $i_n,n\in\{1,\dots,5\}$ now running over $\{0,1,2\}$
\begin{multline}
T_{i_1, i_2,i_3,i_4, i_5}=\\ \sum_{\{\alpha\}}A_{i_1,\alpha_1}^{(1)}A_{\alpha_1, i_2,\alpha_2}^{(2)} A_{\alpha_2, i_3,\alpha_3}^{(3)} A_{\alpha_3, i_4,\alpha_4}^{(4)}A_{\alpha_4,i_5}^{(5)}\,.
\end{multline}

\subsection{Prediction with the Qutrit Model}
The procedure outlined in \Cref{sec:tnprediction} mostly carries over to this new model, as the design is dimension-agnostic. The only changes are the basis vectors for projection and marginalisation being changed to work with the qutrit model. The three basis vectors for $\mathbf{e}^{(i_k)}$ are:
\begin{align*}
\mathbf{e}_{\text{incorrect}}=\ket{0}=\begin{pmatrix}1 &0&0\end{pmatrix}^T,\\ \mathbf{e}_{\text{out-of-order}}=\ket1=\begin{pmatrix}0&1&0\end{pmatrix}^T,\\ \mathbf{e}_\text{in-order}=\ket2=\begin{pmatrix}0 &0&1\end{pmatrix}^T.
\end{align*}
To contract the physical indices of the future sites, we use vector $\mathbf v$:
\begin{equation}
\mathbf v=\begin{pmatrix}1 &1&1\end{pmatrix}^T.
\end{equation}
We then use \Cref{eq:matscoring,eq:argmaxselection} to find which outcome is most likely. The only change being that $s_t$ now runs over $\{0,1,2\}$. For completeness, \Cref{eq:argmaxselection} is now written as:
$$\hat{s}_t=\operatorname*{argmax}_{s_t\in\{0,1,2\}}|R_{s_t}|\,.$$

\section{Encoding Valence in a Tensor Network}\label{sec:valence}

In the previous two Sections, we have demonstrated how tensor network based approaches show suitability for modelling order effects in human memory. Both the initial model, where the experimental results were simply represented with a qubit with correct/incorrect states, and in a more involved qutrit based model with states for recall being completely incorrect, the item being recalled correctly but in the wrong position, and for both the item and position being recalled correctly. They can successfully capture temporal effects such as the primacy and recency effects, by representing the dependencies between past and future recalls through bonds with dimension greater than 1 (which, were this a physical partitioned quantum system, would correspond to entanglement). However, these proposed models construct probability landscapes based entirely on the observed recall outcomes, ignoring the most important part of the study: the emotional valence of each item shown to the participants.

As discussed in the introduction, cognitive literature has shown that human memory is not solely determined by the order of experiences. The emotional valence of experiences significantly influences both the storage and the recollection of the memories formed. A purely outcome-based approach limits a model's capability to express this dependence on emotions. By introducing valence into our tensor network, we could increase the accuracy, or at least determine whether valence can be effectively modelled using similar techniques.

\subsection{Modelling a Trial with Valence}
We treat the experience of seeing an object of a certain valence as two distinct stages, the first encoding the valence which acts as a `priming' stage. This will shift the memory state into a certain configuration based on whether the item was positive or neutral, representing how the brain stores memories differently based the emotion of the experienced event. Then the actual act of viewing the object is encoded as previous.

We reuse our recall label space defined in \Cref{sec:trialqutrit}, referred to now as $\mathcal H_s\cong\mathbb C^3$, with the basis states
\begin{align*}
	\ket{0}_i & \to\text{Incorrect item and position for item }i,      \\
	\ket{1}_i & \to\text{Correct item, incorrect position for item }i, \\
	\ket{2}_i & \to\text{Correct item and position for item }i.
\end{align*}
To encode the valence, we define a two-dimensional label space $\mathcal H_v\cong\mathbb{C}^2$, with the basis states
\begin{align*}
	\ket0_i   & \to\text{Item } i \text{ has neutral valence},         \\
	\ket1_i   & \to\text{Item } i\text{ has positive valence}.
\end{align*}
The tensor product of these two spaces gives us every combination of recall outcome, forming a six-dimensional space $\mathcal H_\text{total}=\mathcal H_v \otimes \mathcal H_s\cong\mathbb{C}^{2\times3}=\mathbb{C}^6$. The six basis vectors of this space are the pairs between each valence, and each recall outcome, written as $\ket{v_i,s_i}$. For item $i$, they are as follows:
\begin{align*}
	\ket{00}_i & \to\text{Neutral, incorrect item and position},       \\
	\ket{01}_i & \to\text{Neutral, correct item, incorrect position},  \\
	\ket{02}_i & \to\text{Neutral, correct item and position},         \\
	\ket{10}_i & \to\text{Positive, incorrect item and position},      \\
	\ket{11}_i & \to\text{Positive, correct item, incorrect position}, \\
	\ket{12}_i & \to\text{Positive, correct item and position}.
\end{align*}
Using the previous method of constructing the global probability tensor, we would write this system as a 10th-order tensor with each valence priming and recall stage separate like so: $$T_{v_0,s_0,v_1,s_1,v_2,s_2,v_3,s_3,v_4,s_4,v_5,s_5},$$ which would form an alternating, 10-node chain $V^{(1)}\to S^{(1)}\to\dots\to V^{(5)}\to S^{(5)}$. This structure would imply that the emotional priming and recall state are distinct, sequential events in time. However, in cognitive psychology, the environmental context and behavioural outcome (in this case, the ability to recall), are generally understood to be inextricably linked~\cite{tulving1983elements, diana2007imaging, kensinger2009remembering}, and this structure cannot enforce this constraint, allowing for uncorrelated valence-recall pairs. By flattening the individual valence and recall indices into a single physical index (referred to as an effective site), where $\ket{e_i}=\ket{v_i}\otimes\ket{s_i}$, we force the tensor network to treat the priming and recall as a single memory event
Using Penrose diagrams, we can visualise how each valence-recall pair is combined:
\begin{figure*}
	\centering
	\begin{subfigure}{\linewidth}
    \captionsetup{ justification=raggedright, singlelinecheck=false }
    \caption{}\label{fig:valence_model}
		\includegraphics[width=\textwidth]{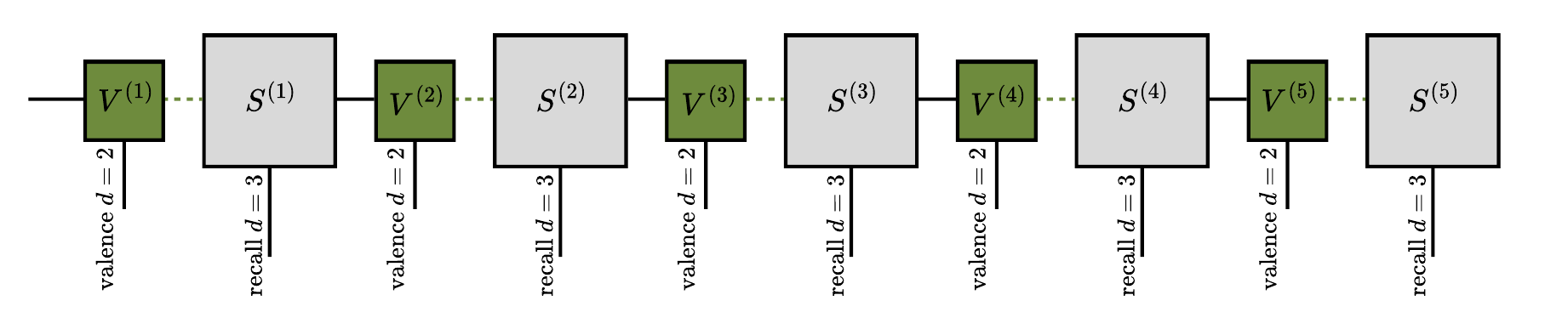}
	\end{subfigure}
	\begin{subfigure}{\linewidth}
    \captionsetup{ justification=raggedright, singlelinecheck=false }
    \caption{}\label{fig:valence_model_effec_site}
		\includegraphics[width=\textwidth]{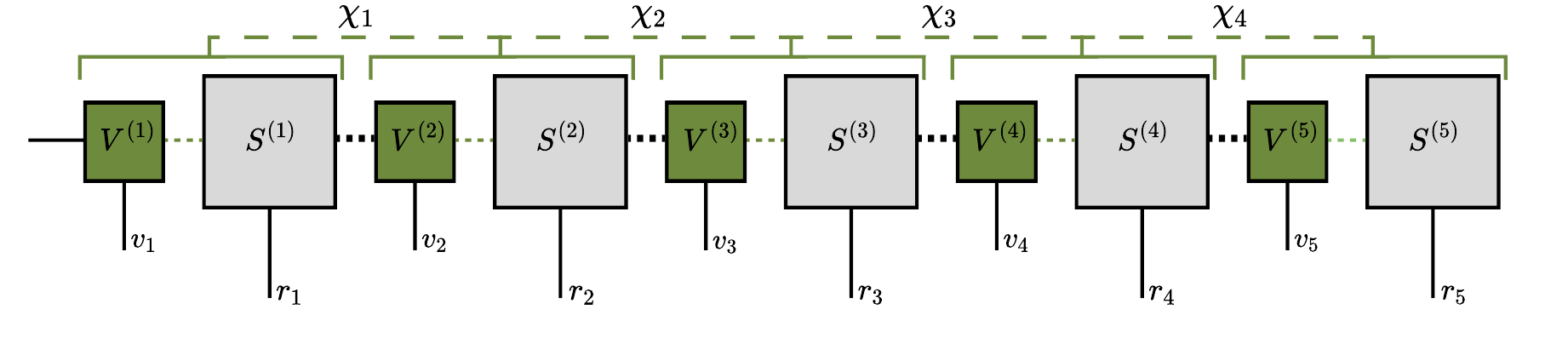}
	\end{subfigure}
    \begin{subfigure}{\linewidth}
    \captionsetup{ justification=raggedright, singlelinecheck=false }
    \caption{}\label{fig:valmps}
	\includegraphics[width=0.75\textwidth]{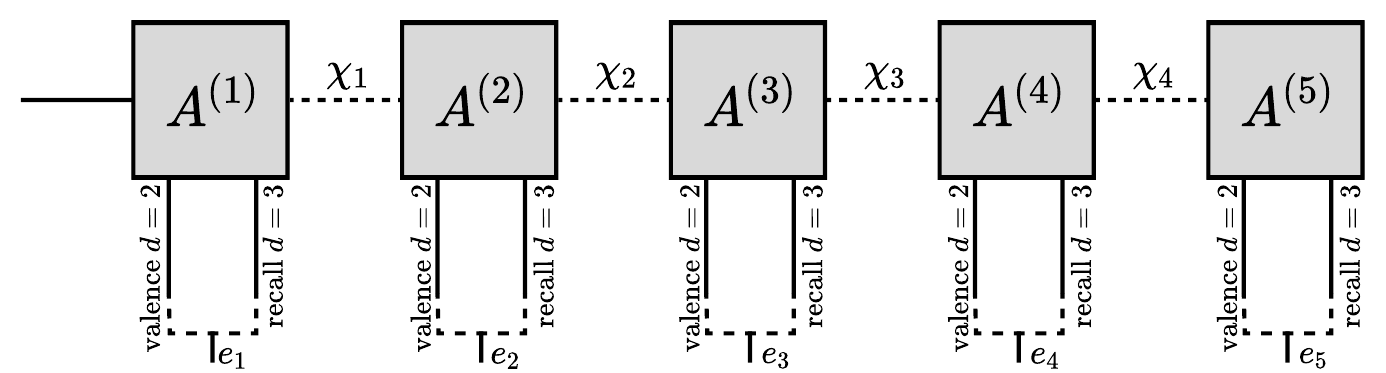}
    \end{subfigure}
	\caption{Penrose diagrams visualising the tensor network representation of combined valence-recall pairs. \Cref{fig:valence_model} shows the baseline sequence of alternating valence ($V$) and recall ($S$) tensors with respective physical dimensions. \Cref{fig:valence_model_effec_site} shows the grouped tensor structure with the discrete physical indices ($v_i,s_i$) and bond dimensions $\chi_i$ connecting each memory event. \Cref{fig:valmps} shows the resultant condensed network, with each pair contracted into a single composite tensor $A^{(i)}$. The distinct physical indices have been combined into a single physical index $e_i$.} \label{fig:valence_model_transformation}
\end{figure*}
This composite-site design represents a departure from all existing quantum cognition models (e.g., ~\cite{busemeyer_quantum_2012,pothos_quantum_2022,busemeyer_incorporating_2025}), which do not consider valence. Valence is not represented as a variable in any existing MPS or projection-operator framework for memory. By encoding valence directly into the physical index of each site, the tensor network is forced to learn the joint distribution over emotional context and recall outcome simultaneously, rather than treating them as separable.
To map from a given pair of physical indices $(v_i,s_i)\to e_i$, we use the following lexicographical mapping:
$$e_i=v_i\times d_s+s_i,$$
where $d_s$ is the dimension of the recall space (3).
We substitute the flattened index $e_i$ into the tensor to construct a 5th-order tensor $T$ which we then use to fill the global probability landscape, with each element of the tensor denoted by:
$$T_{e_1,e_2,e_3,e_4,e_5},$$
where $e_i\in\{0,1,\dots,5\}$ for all $i\in\{1,2,\dots,5\}$.

As our mapping of physical indices to $e_i$ is bijective, derived directly from the tensor product, no information is lost. However, when we perform SVD, and split between $e_n$ and $e_{n+1}$, it is forced to calculate the bond-dimension (``entanglement'') between the composite \textit{valence-recall} states of item $n$ and item $n+1$.

Each index in the probability tensor is defined by
\begin{multline}T_{e_1,e_2,e_3,e_4,e_5}=P(E_1=e_1,E_2=e_2,E_3=e_3,\\E_4=e_4,E_5=e_5),\end{multline}
and is populated from the experimental data as follows:
\begin{equation}T_{e_1, e_2, e_3, e_4, e_5} = \frac{1}{N} \sum_{k=1}^{N}\prod_{m=1}^5\delta_{s_m^{(k)} e_m}
.\label{eq:valencedataload}\end{equation}
$e_m\in\{0,\dots,5\}$ represents the six bipartite states $\ket{v_m,s_m}$ for $m=0,\dots5$.
The following normalisation constraint holds:
$$\sum_{e_1,\dots,e_5}T_{e_1,\dots,e_5}=1.$$

\subsection{Matricisation}
To perform SVD, we must matricise this tensor as we have done previously, changing the base of the exponent to 6 to represent the expanded composite state space. Each resulting matrix does not just represent how the past recall patterns map onto future outcomes, but how the emotional valence of items affects that mapping.

\subsection{Valence Model Matrix Product State}
The MPS for this model with a valence priming stage is constructed in a similar way as outlined in \Cref{sec:ssvd}, using Sequential Singular Value Decomposition. However, each cut is effectively performed every two nodes (after each recall node). Collapsing each six-dimensional site creates an MPS where the bond-dimension (``entanglement'') encodes not just sequence dependencies, but also valence dependencies.
As we have already collapsed each valence-recall pair by flatting their physical indices, we are able to perform the same process given previously. This results in the factorisation below, with the effective index $e_m,m\in\{1,2,\dots,5\}$ running over $\{0,1,\dots,5\}$:
\begin{multline}
	T_{e_1,e_2,e_3,e_4,e_5}=\\\sum_{\{\alpha\}}A_{e_1,\alpha_1}^{(1)}A_{\alpha_1,e_2,\alpha_2}^{(2)}A_{\alpha_2,e_3,\alpha_3}^{(3)}A_{\alpha_3,e_4,\alpha_4}^{(4)}A_{\alpha_4,e_5}^{(5)}.
\end{multline}
In \Cref{fig:valmps}, each tensor $A^{(i)}$ represents a single episodic event, where the distinct valence and recall indices have been collapsed into an effective site. The virtual bonds $\chi$ represent the correlations (or, were this a physical rather than psychological system, entanglement) between each valence-recall event.

\subsection{Prediction}
In the models discussed in \Cref{sec:tnprediction,sec:qutrittn}, predicting the outcome of a specific target item $t\in\{1,\dots,5\}$ was achieved by conditioning the tensor network on past recall observations, and then marginalising all future outcomes. The introduction of a valence priming stage requires a change to this prediction step. As the emotional context (positive or neutral) is `known' to the subject before they attempt to recall, we must also condition the state on the known valence $v_t$ before predicting the recall outcome $s_t$.

\textbf{Projection of Past Observations:}
The six basis vectors for $\mathbf{e}^{(e_k)}$ are:
\begin{align*}
	\intertext{Neutral, incorrect:}
	\ket{0,0}=\mathbf{e}_1 & =\begin{pmatrix}1 & 0 & 0 & 0 & 0 & 0\end{pmatrix}^T
	\intertext{Neutral, correct item, incorrect position:}
	\ket{0,1}=\mathbf{e}_2 & =\begin{pmatrix}0 & 1 & 0 & 0 & 0 & 0\end{pmatrix}^T
	\intertext{Neutral, correct:}
	\ket{0,2}=\mathbf{e}_3 & =\begin{pmatrix}0 & 0 & 1 & 0 & 0 & 0\end{pmatrix}^T
	\intertext{Positive, incorrect:}
	\ket{1,0}=\mathbf{e}_4 & =\begin{pmatrix}0 & 0 & 0 & 1 & 0 & 0\end{pmatrix}^T
	\intertext{Positive, correct item, incorrect position:}
	\ket{1,1}=\mathbf{e}_5 & =\begin{pmatrix}0 & 0 & 0 & 0 & 1 & 0\end{pmatrix}^T
	\intertext{Positive, correct:}
	\ket{1,2}=\mathbf{e}_6 & =\begin{pmatrix}0 & 0 & 0 & 0 & 0 & 1\end{pmatrix}^T
\end{align*}
The physical indices for items in the past $k<t$ of the corresponding tensors $A^{k}$ are pinned by taking the inner product as shown in \Cref{eq:projection}, producing the same transfer matrices $B^{(k)}$ but mapped to the larger label space $\mathcal H_\text{total}$.

\textbf{Marginalisation of Future Outcomes:}
For the sites $k>t$, they are marginalised out. The physical indices of the future tensors are contracted with a 6-dimensional vector of ones:
\begin{equation}
\mathbf{v}=\begin{pmatrix}1&1&1&1&1&1\end{pmatrix}^{T},
\end{equation}
to create the transfer matrices $B^{(k)}$ following \Cref{eq:conditioning}.

\textbf{Conditioning the Target Tensor on Valence:}
The main difference from the previous two models is how the target tensor $A^{(t)}$ is treated. The index of the physical site $e_t$ is represented by the composite state $\ket{v_t,s_t}$. A naive prediction would marginalise over both valence and recall outcomes at site t. As the valence is already known, we don't evaluate the probability over the whole 6-dimensional space. Instead, we condition the target tensor on $v_t$, which projects it into the relevant 3-dimensional recall subspace, reflecting how the experiment actually happens. In existing quantum cognition models, all observables are treated symmetrically: asking question A before B changes the answer via non-commutativity, but there is no mechanism to inject external context into the measurement process. Conditioning on valence allows us to model real experiences more accurately, as the emotional character of an item is not a question asked about the item. It is an inherent property `known' before recall is attempted.

If the item $t$ has a neutral valence ($v_t=0$), the valid effective site indices are limited to $e_t\in\{0,1,2\}$. If the item $t$ has a positive valence $v_t=1$, the valid effective site indices are limited to $e_t\in\{3,4,5\}$. Formally, for a given known valence $v_t$, the only effective indices with non-zero probabilities are:
$$\mathcal E_{v_t} = \{v_t\times 3+s_t:s_t\in{0,1,2}\}.$$
This leaves the target tensor $A^{(t)}$ partially open, with its physical index $e_t$ restricted to the subset of its known valence.

The left environment $E_L$ and the right environment $E_R$ are constructed following \Cref{eq:leftenv,eq:rightenv}. The score for each outcome $e_t$ is then computed as:
$$R(e_t)=E_L A^{(t)}(e_t)E_R,\quad e_t\in\mathcal E_{v_t},$$
where $A^{(t)}(e_t)$ denotes the two-dimensional matrix obtained by selecting the slice of the rank-3 tensor $A^{(t)}$ at the physical index $e_t$.
The prediction is then made by choosing the state that is most likely:
$$\hat e_t =\operatorname*{argmax}_{e_t\in\mathcal E_{v_t}}|R(e_t)|,$$
and recovering the predicted recall outcome from $\hat e_t$ by inverting the lexicographical mapping:
\begin{align*}\hat s_t&=\hat{e_t}-v_t\times d_s\\&=\hat e_t\mod3,\end{align*}
yielding $s_t\in\{0,1,2\}$, mapping to the three recall outcomes. Restricting the scoring to only the subspaces of $\mathcal H_{total}$ that are valid for the known valence of $t$ means the model never assigns a probability to a recall outcome with the incorrect valence.

\section{Tensor Network Data Analysis}\label{sec:TNDataAnalysis}

This section assesses how effective tensor networks are when modelling human memory. The methodology presented here allows us to determine how well each of the three tensor network models capture the dependencies apparent in memory, and their ability to predict recall states across a sequence. The three models that have been discussed are:
\begin{enumerate}
	\item An initial `toy' tensor network, which just models outcomes as forgotten/remembered.
	\item An expanded tensor network which introduces a third outcome (remembered but in the incorrect order)
	\item A tensor network with valence, which adds a priming step based on the item's emotional valence.
\end{enumerate}
For each model, a cross-validation process was used to prevent overfitting. Prediction accuracy was calculated for all individual sequence positions, and for the overall sequence. Individual recall classes were also examined, to see whether models are better at predicting certain recall outcomes. Scree plots were produced for each bond in each MPS, showing how the values decay inside the diagonal matrix $\bm{\Sigma}$ produced by SVD. These plots show the internal structure of each MPS, and are used to determine whether a model is successfully modelling order and valence dependencies, or just memorising states. Additionally, the SVD truncation threshold was swept through a range of values to see how the models perform when compressed. Doing this allows us to determine how much accuracy is lost when constructing smaller and more compact networks, to find a balance between effectiveness at modelling memory and computational efficiency.

\subsection{Validation of Use of Tensor Networks}

Before we analyse the individual models, it is worth examining the raw structure of the experimental data itself, as this provides the basis upon which all three models are built, and motivates why a tensor network approach is appropriate.

\begin{figure}
	\begin{center}
    \begin{subfigure}{\linewidth}
    \captionsetup{ justification=raggedright, singlelinecheck=false }
    \caption{}\label{fig:qubitsurface}
		\includegraphics[width=\textwidth, trim=6cm 0cm 0cm 0cm]{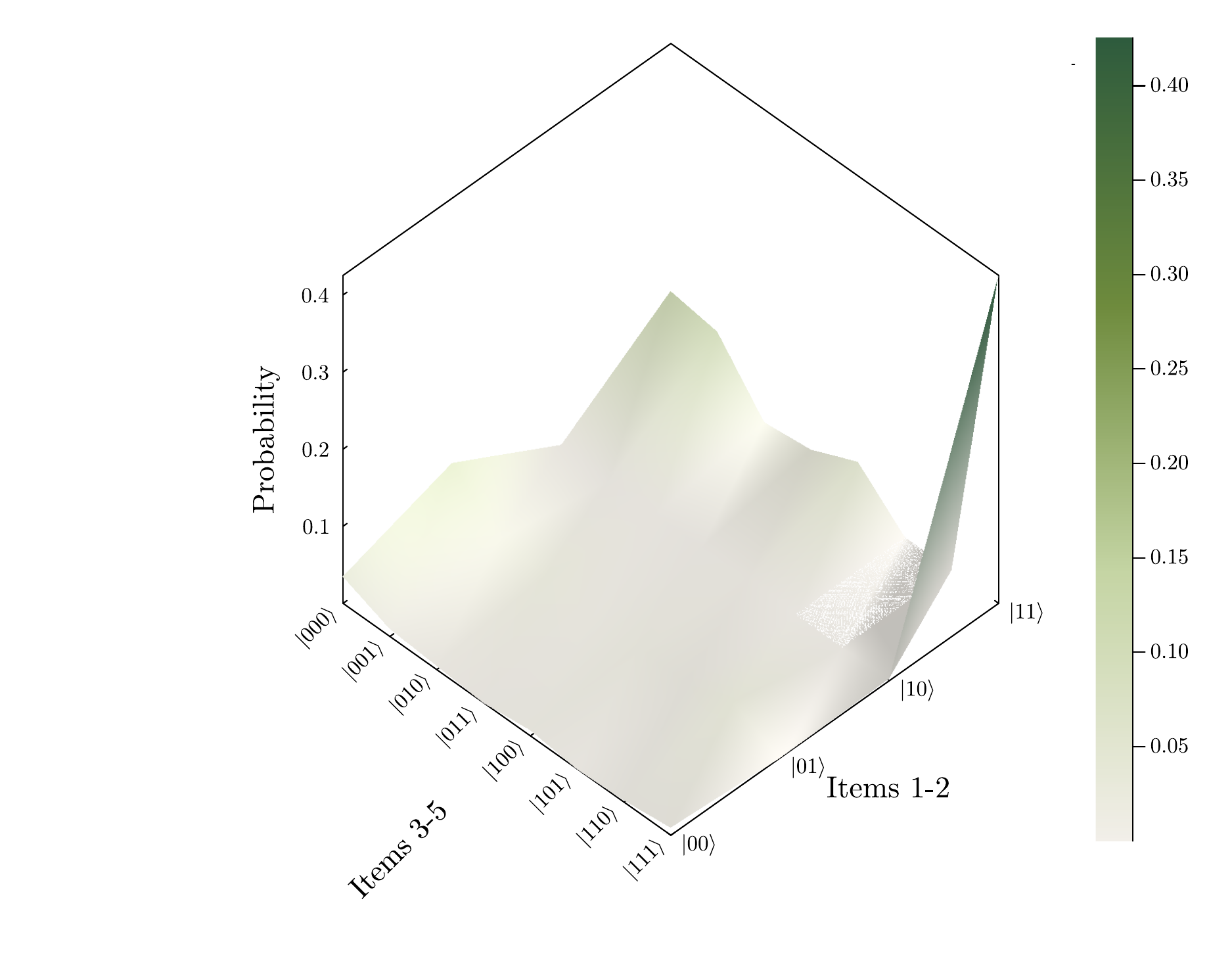}
	\end{subfigure}
     \begin{subfigure}{\linewidth}
    \captionsetup{ justification=raggedright, singlelinecheck=false }
    \caption{}\label{fig:qutritsurface}
		\includegraphics[width=\linewidth, trim=6cm 0cm 0cm 0cm ]{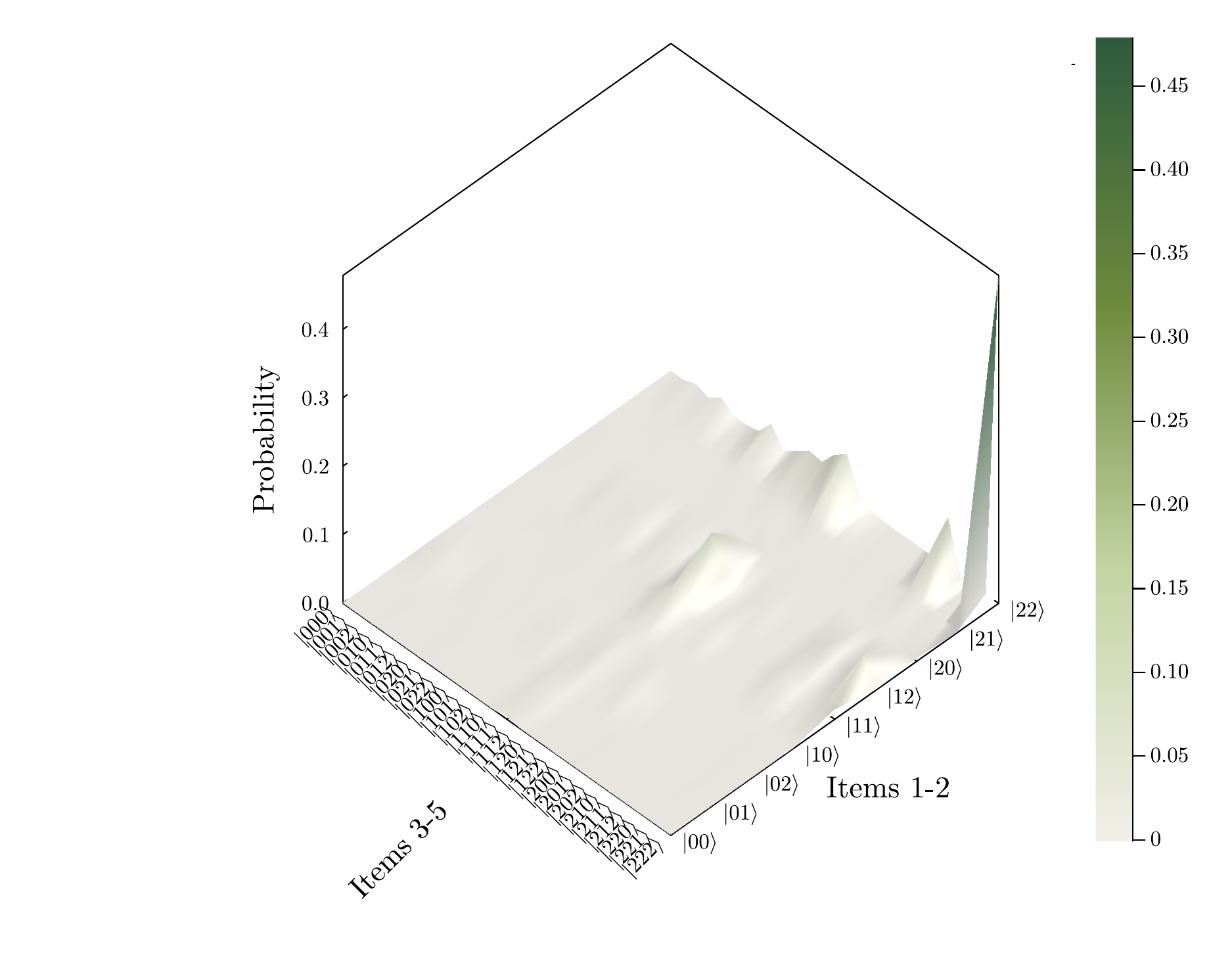}
	\end{subfigure}
	\end{center}
	\caption{Qubit (\Cref{fig:qubitsurface}) and qutrit (\Cref{fig:qutritsurface}) probability tensor surface visualisation, $k=2$, $N=404$.}\label{fig:qubitqutritsurface}
\end{figure}

\Cref{fig:qubitsurface,fig:qutritsurface} show the surface plots of the matricised joint probability tensor constructed in \Cref{sec:3matricisation,sec:4matricisation}, after being split at $k=2$. The matrix encodes the joint probability of every combination of recall outcomes for items 1–2 (rows) against items 3–5 (columns). If recall of each item were independent, the frequencies would be spread uniformly across all cells.

In the qubit surface plot \Cref{fig:qubitsurface}, the highest peak represents a participant recalling the first two items correct ($\ket{11}$), and then recalling the 3 items after the cut correct ($\ket{111}$). This shows that a significant number (almost half) of trials, participants had perfect recall. There is a variety of other recall patterns after $\ket{11}$, ranging from forgetting all items that came after to remembering most of them. There is also a noticeable number of participants who appear to have only remembered one or two items, however this is most likely not the case, and is in fact due to the` limited outcomes of this model, where items that were remembered but just in the wrong order are counted as an incorrect recall.

This plot visually shows why tensor networks are necessary for this data. If memory recall were a simple or independent process, this surface would look like a smooth, gentle hill. Instead, it forms a jagged landscape with specific concentrations of values. Matrix Product States were designed specifically to represent these `spiky', highly concentrated state spaces, and by breaking down the sequence into local tensors, an MPS efficiently maps the sequential correlations (which, for a physical rather than psychological system, would be considered ``entanglement'') between items. It can represent the sharp peak of perfect recall whilst also representing the areas of partial recall without wasting computational resources or parameters on the empty valleys where probability is near zero.

The qutrit surface plot (\Cref{fig:qutritsurface}) show the same results, just in a larger state space. The perfect recall peak still dominates results (as would be expected), but there is significantly more detail in the structure. The large number of fully incorrect responses has disappeared, as those are now represented by the new outcome introduced in this mode, where they remembered the item but in the wrong order. There are many smaller peaks along the $\ket{22}$ row of the Items 1–2 axis. This suggests that when participants recalled the first two items correctly and in order, there was meaningful variation in how they recalled the remaining items. Some were out-of-order, some were completely forgotten. The qutrit representation captures these distinctions that the qubit model ignores.

\begin{figure*}
	\begin{center}
		\includegraphics[width=0.9\textwidth]{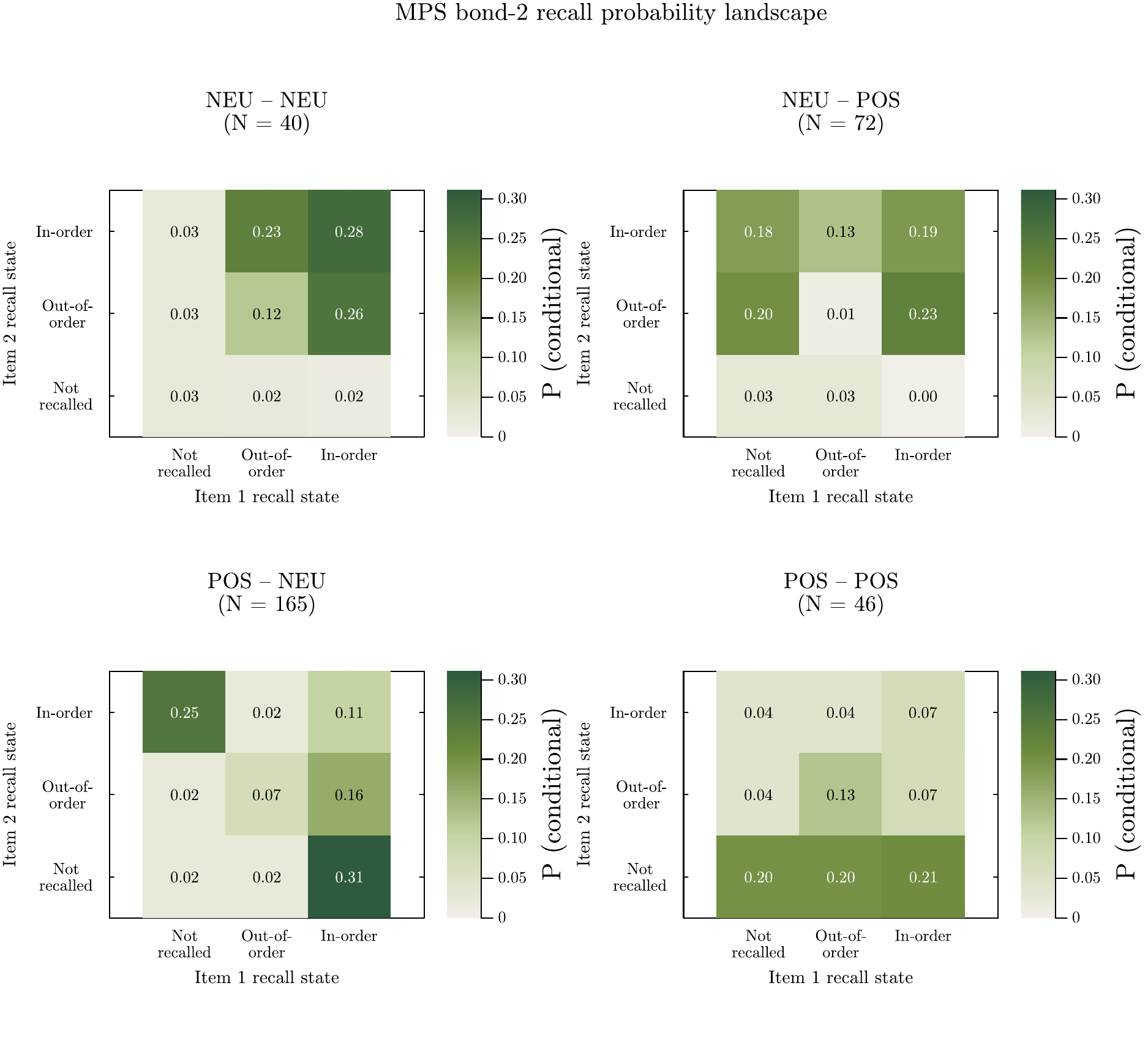}
	\end{center}
	\caption{Probability of recall of Item 2 in a sequence, based on the recall state and valence of Item 1 in that sequence, and the valence of that Item 2.
    }\label{fig:bond2_heatmap}
\end{figure*}

\textbf{Valence Priming in Tensor Networks:} \Cref{fig:bond2_heatmap} shows the probability landscape of bond 2 of the valence-primed model. Each subplot represents a different combination of valences for the first two items. The four quadrants are structurally distinct, providing direct empirical evidence that emotional valence reshapes the joint recall distribution and cannot be treated as an independent factor.

In the NEU-NEU quadrant, the probabilities are concentrated in the top right corner, with both items being recalled in-order being the highest probability, and one item being recalled in-order and the other out-of-order being the second highest. This suggests that when both items are neutral, participants successfully remembered both items, though sometimes mixed their order up.

The NEU-POS quadrant shows that a positive item 2 tends to be recalled successfully, regardless of how item 1 was recalled, consistent with the emotional memory enhancement effect~\cite{anderson_emotional_2006,bennion_oversimplification_2013}: positive valence strengthens encoding of the second item.

The POS-NEU quadrant shows that participants who correctly recalled the positive item were more likely to fail to remember the second item than they were to recall it both in-order and out-of-order. This pattern, where an emotional item appears to disrupt recall of the subsequent neutral item, aligns with the central/peripheral trade-off discussed in~\cite{lanciano_memory_2011, payne_impact_2006}. The attentional resources drawn by the positive item appear to come at the cost of encoding the item that follows it.

With the POS-POS pattern, there are uniform values across the bottom row, which implies that if both items are positive, the first doesn't seem to influence the recall outcome of the second. The second positive item is just not recalled well generally when following a positive item. This is different from the pattern seen in NEU-POS, where the first item being recalled correctly suggests that the following neutral item will not be remembered at all. Two consecutive emotionally positive items appear to create competing encoding demands, where each item draws attentional resources that would otherwise support the encoding of the other.

It is important to note the small sample size of NEU-NEU and POS-POS patterns, so specific cell level analysis should be done carefully. However, the overall structures here are clearly apparent, and a larger dataset should only reinforce this.
Together these four quadrants demonstrate that valence is not merely a scalar modifier of recall probability, but a structural variable that reshapes the entire joint distribution. A model that ignores valence, like the initial qubit and qutrit models, is therefore missing a key dimension of the data. This motivates the valence-primed model, and provides a theoretical grounding for why its performance improves substantially.

\subsection{Predictive Performance}

\begin{figure}[t]
	\begin{center}
     \begin{subfigure}{\linewidth}
    \captionsetup{ justification=raggedright, singlelinecheck=false }
    \caption{Qubit Model}\label{fig:qubitperf}
		\includegraphics[width=\linewidth]{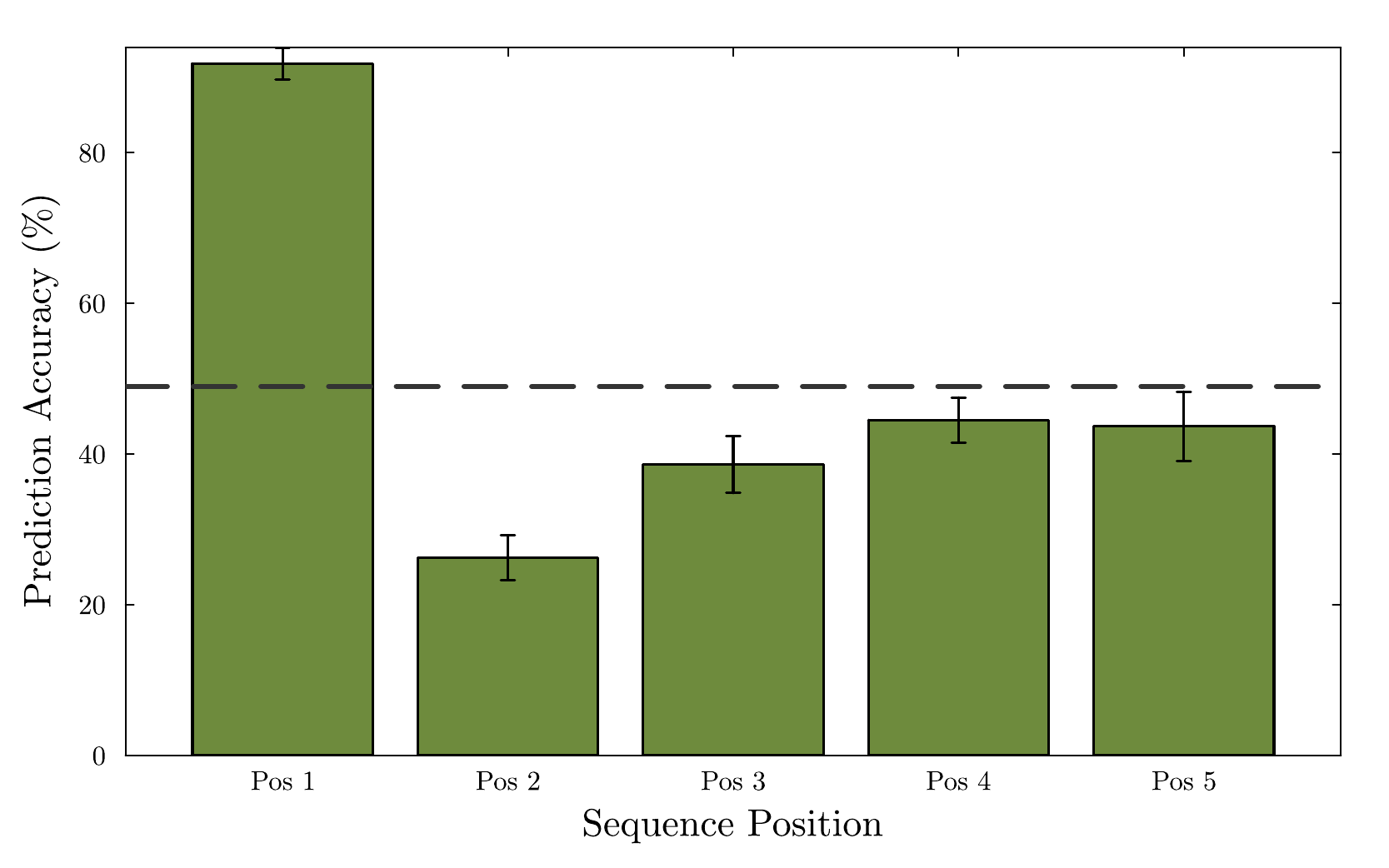}
	\end{subfigure}
    \begin{subfigure}{\linewidth}
    \captionsetup{ justification=raggedright, singlelinecheck=false }
    \caption{Qutrit Model}\label{fig:qutritperf}
		\includegraphics[width=\linewidth]{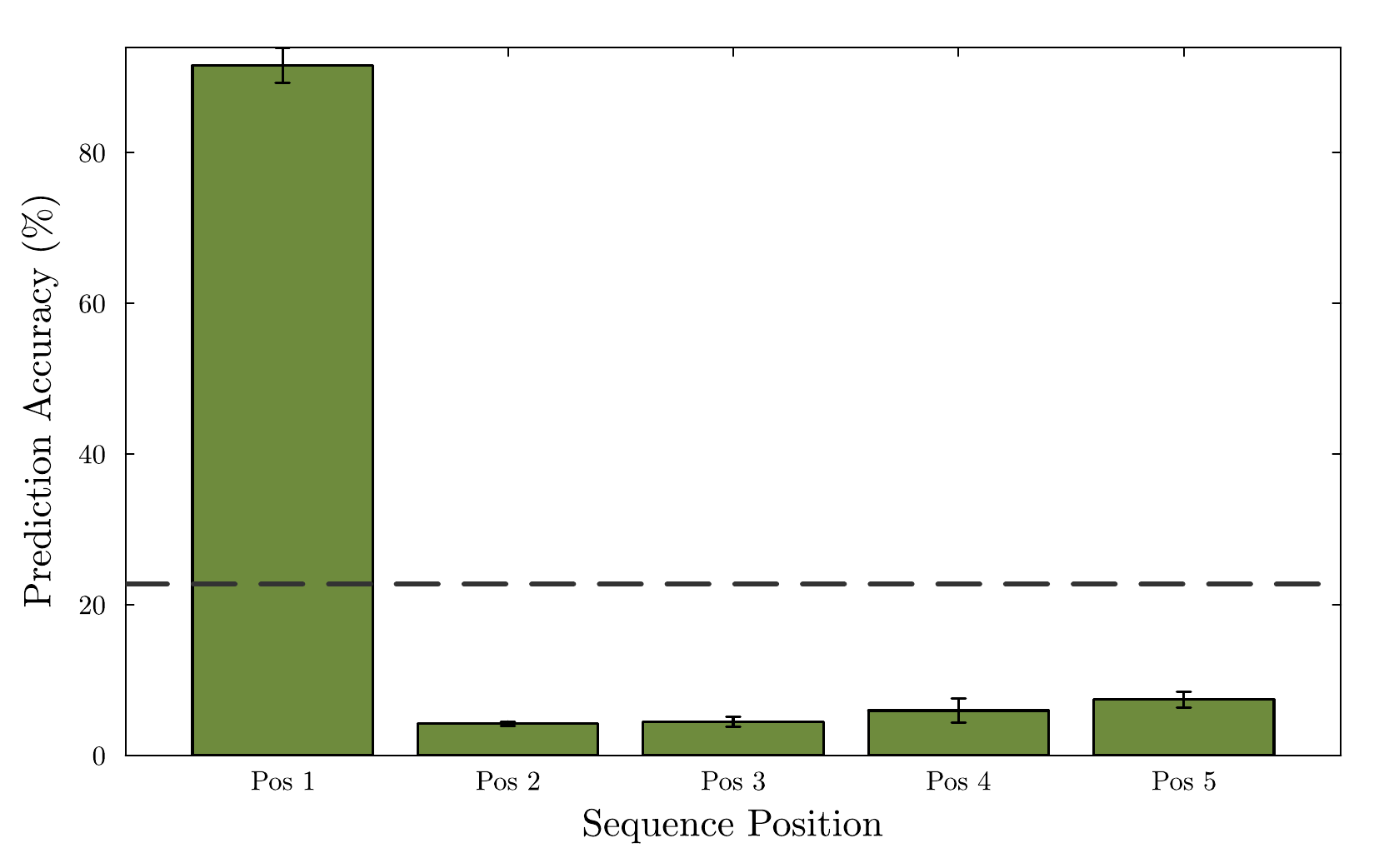}
	\end{subfigure}
		\begin{subfigure}{\linewidth}
    \captionsetup{ justification=raggedright, singlelinecheck=false }
    \caption{Valence-Primed Qutrit Model}\label{fig:primedperf}
		\includegraphics[width=\linewidth]{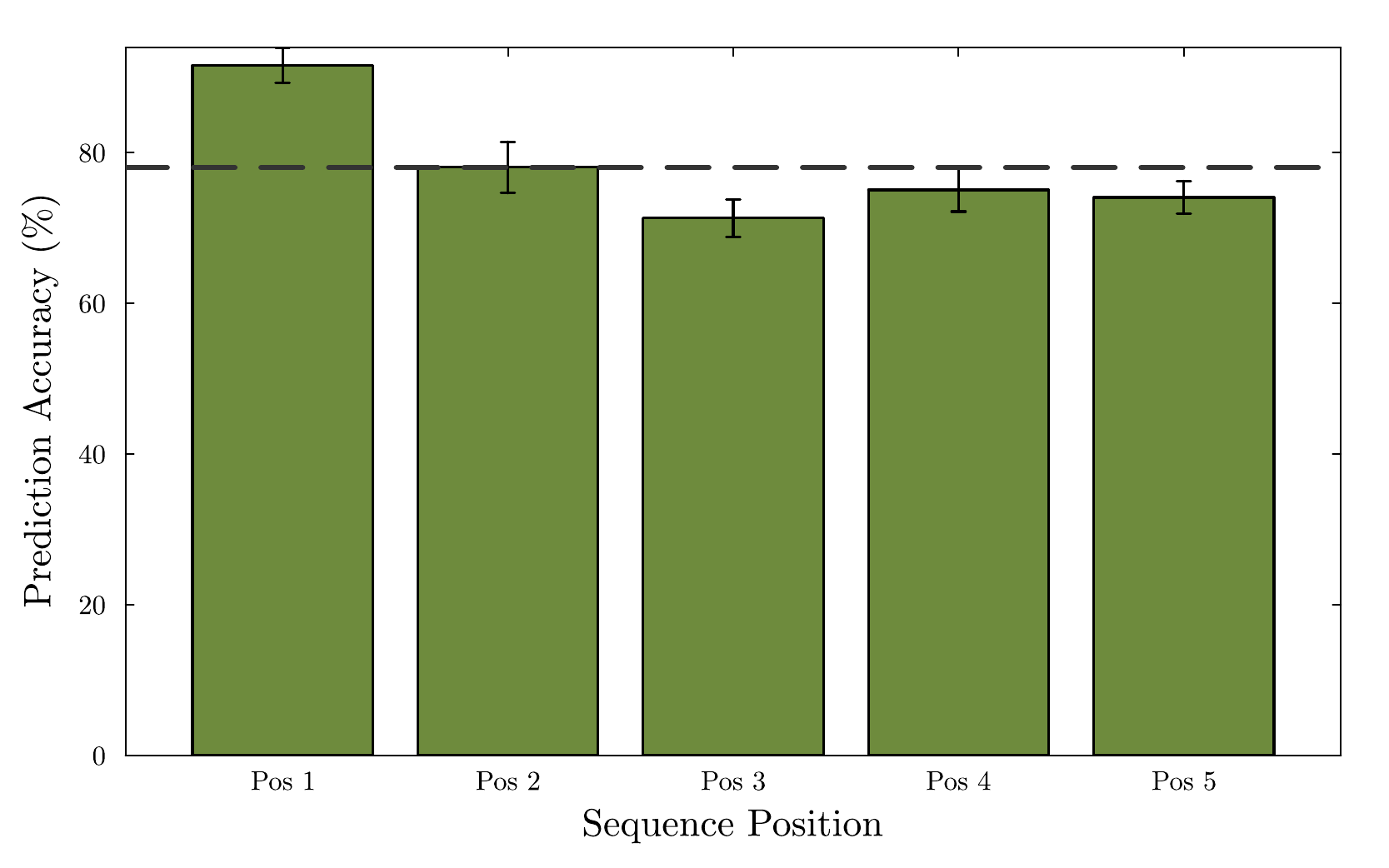}
	\end{subfigure}
	\end{center}
	\caption{Matrix product state cross-validation performance by sequence position for the qubit (\Cref{fig:qubitperf}), qutrit (\Cref{fig:qutritperf}), and valence-primed qutrit (\Cref{fig:primedperf}) models given above.}\label{fig:CrossValPerf}
\end{figure}

\textbf{Qubit Model:} The standard MPS qubit model achieves an overall sequence accuracy of 48.96\% (\Cref{tab:mps_standard}), falling below the 50\% binary chance baseline. Per-class metrics (\Cref{tab:metrics_qubit}) reveal this is driven by severe class imbalance. The model defaults to predicting the majority `incorrect' class (Recall: 0.97, Precision: 0.45), as the initial encoding did not distinguish between positional errors and complete forgetting.

\begin{table}[t]
     \begin{subtable}{\linewidth}\centering
    \captionsetup{ justification=raggedright, singlelinecheck=false }
    \caption{Qubit Model}\label{tab:mps_standard}
	\begin{tabular}{lccc}
		\hline
		\textbf{Position} & \textbf{Mean Acc (\%)} & \textbf{Std Dev} & \textbf{SEM}  \\ \hline
		1             & 91.80                  & 4.69             & 2.10          \\
		2             & 26.23                  & 6.61             & 2.96          \\
		3             & 38.60                  & 8.43             & 3.77          \\
		4             & 44.51                  & 6.66             & 2.98          \\
		5             & 43.67                  & 10.27            & 4.59          \\ \hline
		\textbf{Overall}  & \textbf{48.96}         & \textbf{5.69}    & \textbf{2.55} \\ \hline
	\end{tabular}
\end{subtable}
\begin{subtable}{\linewidth}\centering
    \captionsetup{ justification=raggedright, singlelinecheck=false }
    \caption{Qutrit Model}\label{tab:mps_qutrit}
	\begin{tabular}{lccc}
		\hline
		\textbf{Position} & \textbf{Mean Acc (\%)} & \textbf{Std Dev} & \textbf{SEM}  \\ \hline
		1             & 91.55                  & 5.23             & 2.34          \\
		2             & 4.20                   & 0.63             & 0.28          \\
		3             & 4.45                   & 1.41             & 0.63          \\
		4             & 5.95                   & 3.49             & 1.56          \\
		5             & 7.39                   & 2.30             & 1.03          \\ \hline
		\textbf{Overall}  & \textbf{22.71}         & \textbf{1.46}    & \textbf{0.65} \\ \hline
	\end{tabular}
\end{subtable}
\begin{subtable}{\linewidth}\centering
    \captionsetup{ justification=raggedright, singlelinecheck=false }
    \caption{Valence-Primed Qutrit Model}\label{tab:mps_valence}
	\begin{tabular}{lccc}
		\hline
		\textbf{Position} & \textbf{Mean Acc (\%)} & \textbf{Std Dev} & \textbf{SEM}  \\ \hline
		1             & 91.55                  & 5.23             & 2.34          \\
		2             & 78.27                  & 7.35             & 3.29          \\
		3             & 71.29                  & 5.52             & 2.47          \\
		4             & 75.02                  & 6.42             & 2.87          \\
		5             & 73.79                  & 4.31             & 1.93          \\ \hline
		\textbf{Overall}  & \textbf{77.98}         & \textbf{4.57}    & \textbf{2.04} \\ \hline
	\end{tabular}
\end{subtable}
	\caption{Predictive performance of the qubit (\Cref{tab:mps_standard}), qutrit (\Cref{tab:mps_qutrit}), and valence-primed qutrit (\Cref{tab:mps_valence}) MPS models.}\label{tab:mpspredictiveperformance}
\end{table}

\begin{table}[t]
    \begin{subtable}{\linewidth}\centering
    \captionsetup{ justification=raggedright, singlelinecheck=false }
    \caption{Qubit Model}\label{tab:metrics_qubit}
	\begin{tabular}{lccc}
		\hline
		\textbf{Class} & \textbf{Precision} & \textbf{Recall} & \textbf{F1-Score} \\
		\hline
		Incorrect      & 0.4506             & 0.9733          & 0.6160            \\
		Correct        & 0.9506             & 0.3020          & 0.4583            \\
		\hline
	\end{tabular}
    \end{subtable}
    \begin{subtable}{\linewidth}\centering
    \captionsetup{ justification=raggedright, singlelinecheck=false }
    \caption{Qutrit Model}\label{tab:metricsqutrit}
    	\begin{tabular}{lccc}
		\hline
		\textbf{Class} & \textbf{Precision} & \textbf{Recall} & \textbf{F1-Score} \\
		\hline
		Forgotten      & 0.0617             & 0.9524          & 0.1159            \\
		Out-of-Order   & 0.0000             & 0.0000          & 0.0000            \\
		In-Order       & 0.9506             & 0.3031          & 0.4597            \\
		\hline
	\end{tabular}
    \end{subtable}
    \begin{subtable}{\linewidth}\centering
    \captionsetup{ justification=raggedright, singlelinecheck=false }
    \caption{Valence-Primed Qutrit Model}\label{tab:metricsvalence}
    \begin{tabular}{lccc}
		\hline
		\textbf{Class} & \textbf{Precision} & \textbf{Recall} & \textbf{F1-Score} \\
		\hline
		Forgotten      & 0.1000             & 0.0476          & 0.0645            \\
		Out-of-Order   & 0.7419             & 0.5308          & 0.6188            \\
		In-Order       & 0.7848             & 0.9331          & 0.8525            \\
		\hline
	\end{tabular}
    \end{subtable}
	\caption{Detailed class metrics for the qubit (\Cref{tab:metrics_qubit}), qutrit (\Cref{tab:metricsqutrit}), and valence-primed qutrit (\Cref{tab:metricsvalence}) models.}\label{tab:metricsall}
\end{table}

Position 1 accuracy is very high (\Cref{fig:qubitperf}), reflecting the empirical primacy effect where most participants correctly recall the first item (\Cref{fig:qubitsurface}). Accuracy drops sharply at position 2 before partially recovering towards the sequence end, mirroring expected recency effects~\cite{Glanzer1966Recall}. The accuracy at position 2 falls to 28.4\%, with positions 3-5 recovering to around 40\%.

Taken together, the qubit model demonstrates that a tensor network representation can capture some of the dependency structure in the recall data, as evidenced by the stable CV performance. However, the heavy bias toward the majority class and the low F1 on correct recall indicate that the binary state space is too coarse to represent the subtleties of recall behaviour. Collapsing all recall errors, whether they were just out-of-order or forgotten completely, into a single incorrect state discards information that may be essential for accurate prediction, motivating the expanded qutrit representation.

\textbf{Qutrit Model:} Expanding the state space to distinguish between forgetting, out-of-order recall and in-order recall reduces the overall accuracy to 22.71\% (\Cref{tab:mps_qutrit}). The model fails to predict out-of-order recalls (F1-score: 0.00) and defaults to predicting `forgotten' (\Cref{tab:metricsqutrit}). 

There are two main reasons for this. Firstly, out-of-order recalls are a minority class in the dataset, only making up $\sim 26\%$ of individual item recall results. Secondly, expanding from $d=2$ to $d=3$ increases the state space from 32 to 243 possible recall sequences, significantly increasing data sparsity with the limited number ($N=404$) trials. Many tensor entries are estimated almost entirely from the Laplace prior, and the model cannot reliably learn the conditional distribution of the class.

In \Cref{fig:qutritperf}, item 1 again achieves 91.55\% for the same reason as before. Predictions for positions 2 through 5 are much worse however, falling to between 4.2\% and 7.39\%, well below random chance, with small error bars confirming this is stable across folds. Introducing a third recall state was theoretically motivated, but the available data is simply too sparse to support it. The distinction between positional errors and complete forgetting is a meaningful one cognitively, but 404 trials spread across 243 possible sequences does not appear give the model enough to learn from.

\textbf{Valence-primed qutrit model:} Introducing a valence priming stage before each recall site increases prediction accuracy to 77.98\% (\Cref{tab:mps_valence}). The model is also able to predict the out-of-order class that the qutrit model was unable to (F1-score: 0.618), and correctly predicts the majority of in-order recalls (\Cref{tab:metricsvalence}). This suggests that valence is the specific information enabling the tensor network to distinguish positional errors from complete forgetting. In \Cref{fig:primedperf}, we can see that the position 1 accuracy has remained the same, but all other positions now have a very similar accuracy of $\sim 70\%$. This uniformity suggests that the valence conditioning provides the model with sufficient context to predict recall outcomes consistently, regardless of where an item falls in the sequence. The error bars remain small across all folds, confirming this is a stable result. This model however falls short on predicting items being entirely forgotten. Only 4\% of the recall outcomes across all trials and positions were forgotten, and the model fails to capture any meaningful information from this limited data.
In this specific study, participants were able to recall the items they saw the majority of the time, in-order or not.

\subsection{Singular Value Decay}

\begin{figure}
	\begin{center}
     \begin{subfigure}{\linewidth}
    \captionsetup{ justification=raggedright, singlelinecheck=false }
    \caption{Qubit Model}\label{fig:svdscree}
    \centering
		\includegraphics[width=0.95\linewidth]{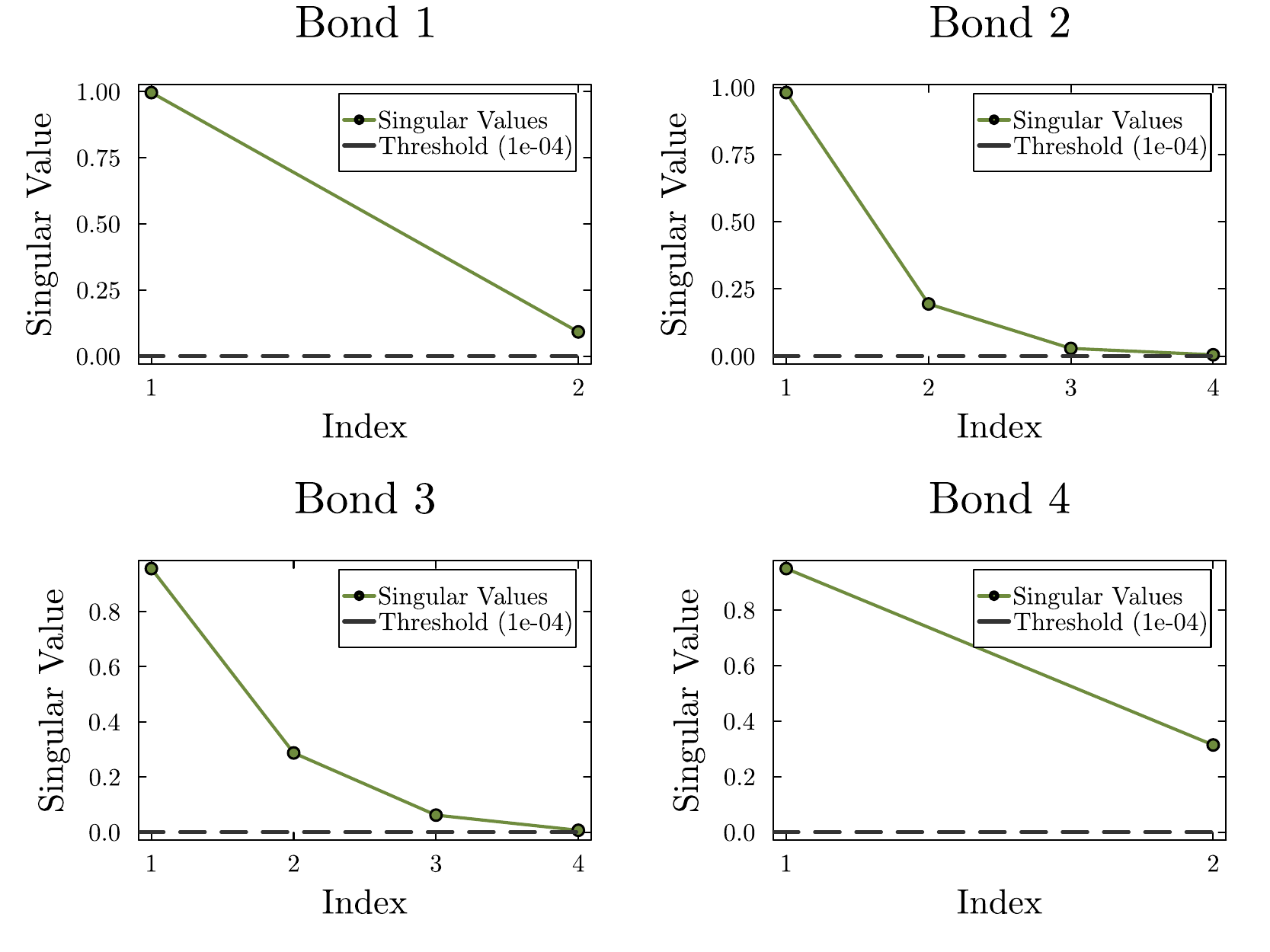}
	\end{subfigure}
    \begin{subfigure}{\linewidth}
    \captionsetup{ justification=raggedright, singlelinecheck=false }
    \caption{Qutrit Model}\label{fig:svdscreequtrit}
    \centering
		\includegraphics[width=0.95\linewidth]{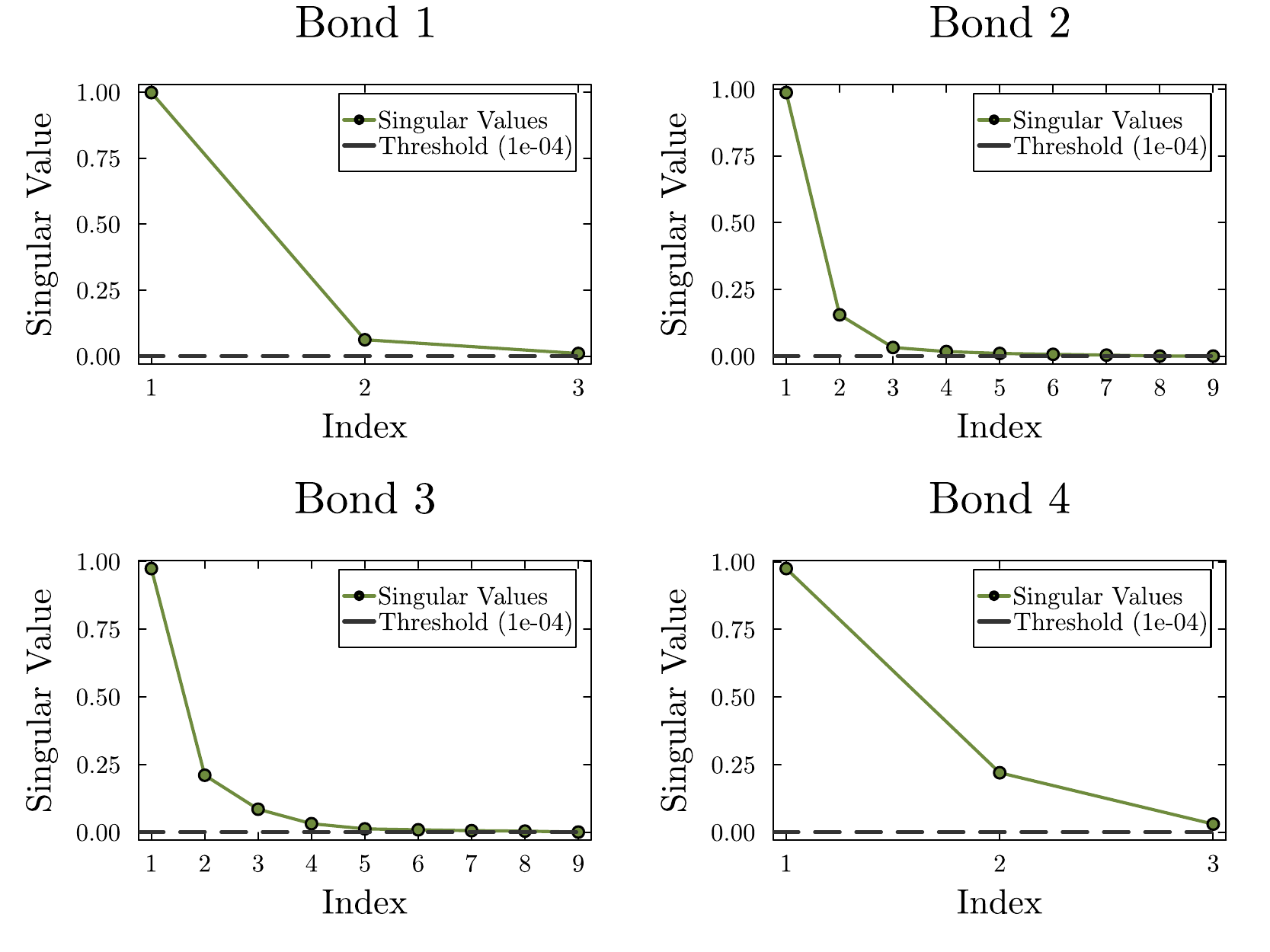}
	\end{subfigure}
		\begin{subfigure}{\linewidth}
    \captionsetup{ justification=raggedright, singlelinecheck=false }
    \caption{Valence-Primed Qutrit Model}\label{fig:svdscreeprimed}
    \centering
		\includegraphics[width=0.95\linewidth]{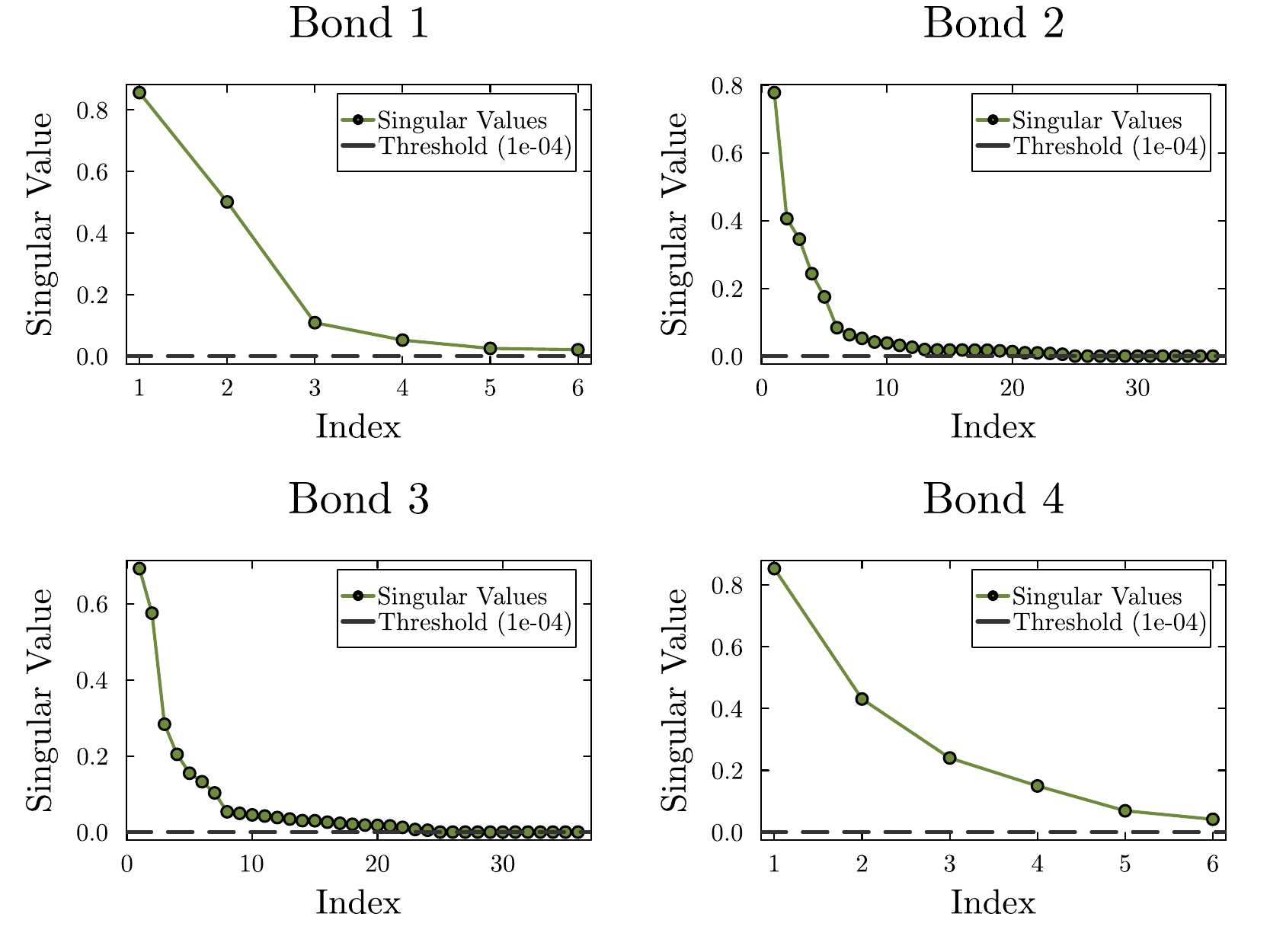}
	\end{subfigure}
	\end{center}
	\caption{Singular value decay for each bond in the qubit (\Cref{fig:qubitperf}), qutrit (\Cref{fig:qutritperf}), and valence-primed qutrit (\Cref{fig:primedperf}) tensor network.}\label{fig:svdscreeall}
\end{figure}

\begin{figure}[t]
	\begin{center}
    \begin{subfigure}{\linewidth}
    \captionsetup{ justification=raggedright, singlelinecheck=false }
    \caption{Qubit Model}\label{fig:qubitsweep}
		\includegraphics[width=\linewidth]{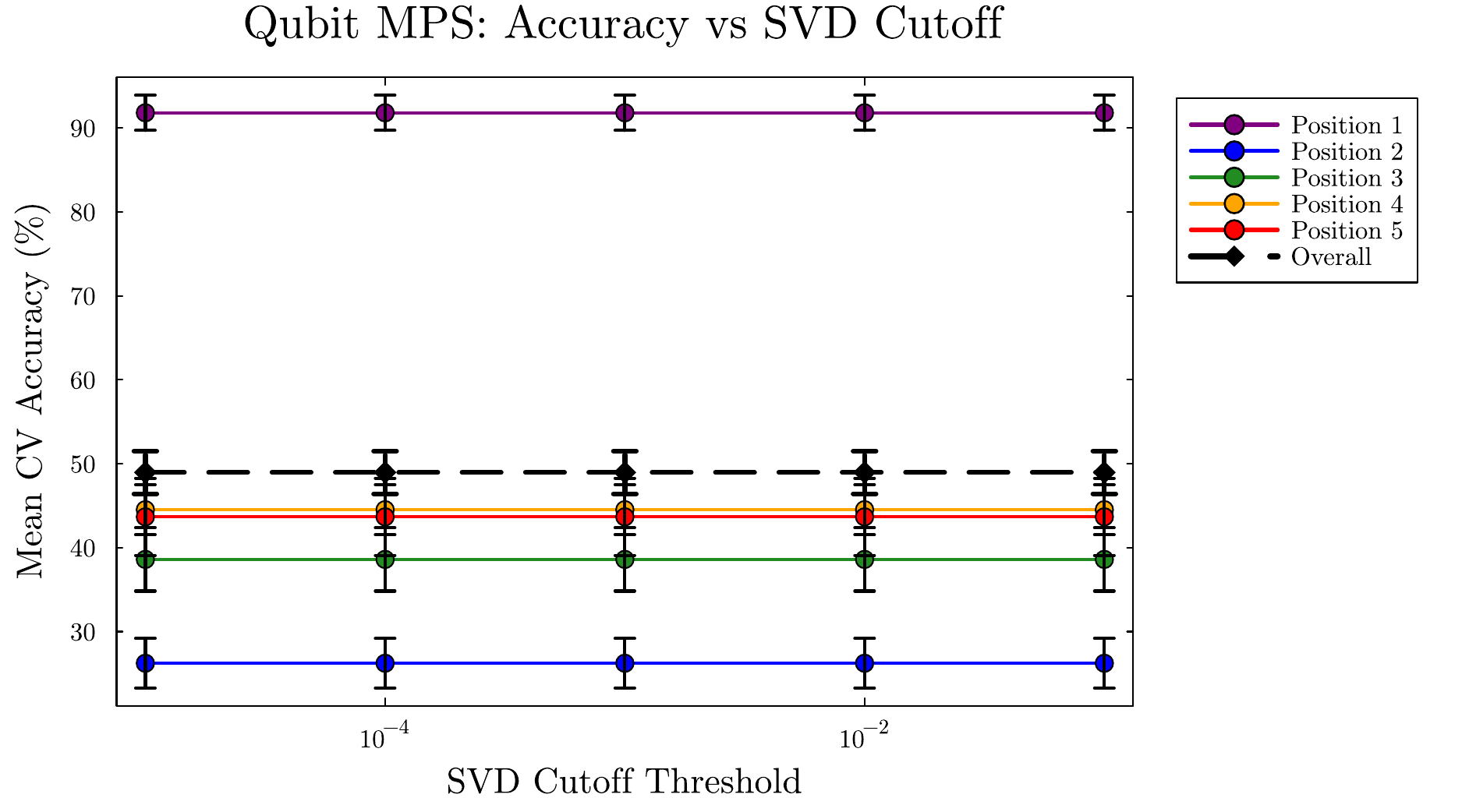}
	\end{subfigure}
	\begin{subfigure}{\linewidth}
    \captionsetup{ justification=raggedright, singlelinecheck=false }
    \caption{Valence-primed Qutrit Model}\label{fig:primedsweep}
		\includegraphics[width=\linewidth]{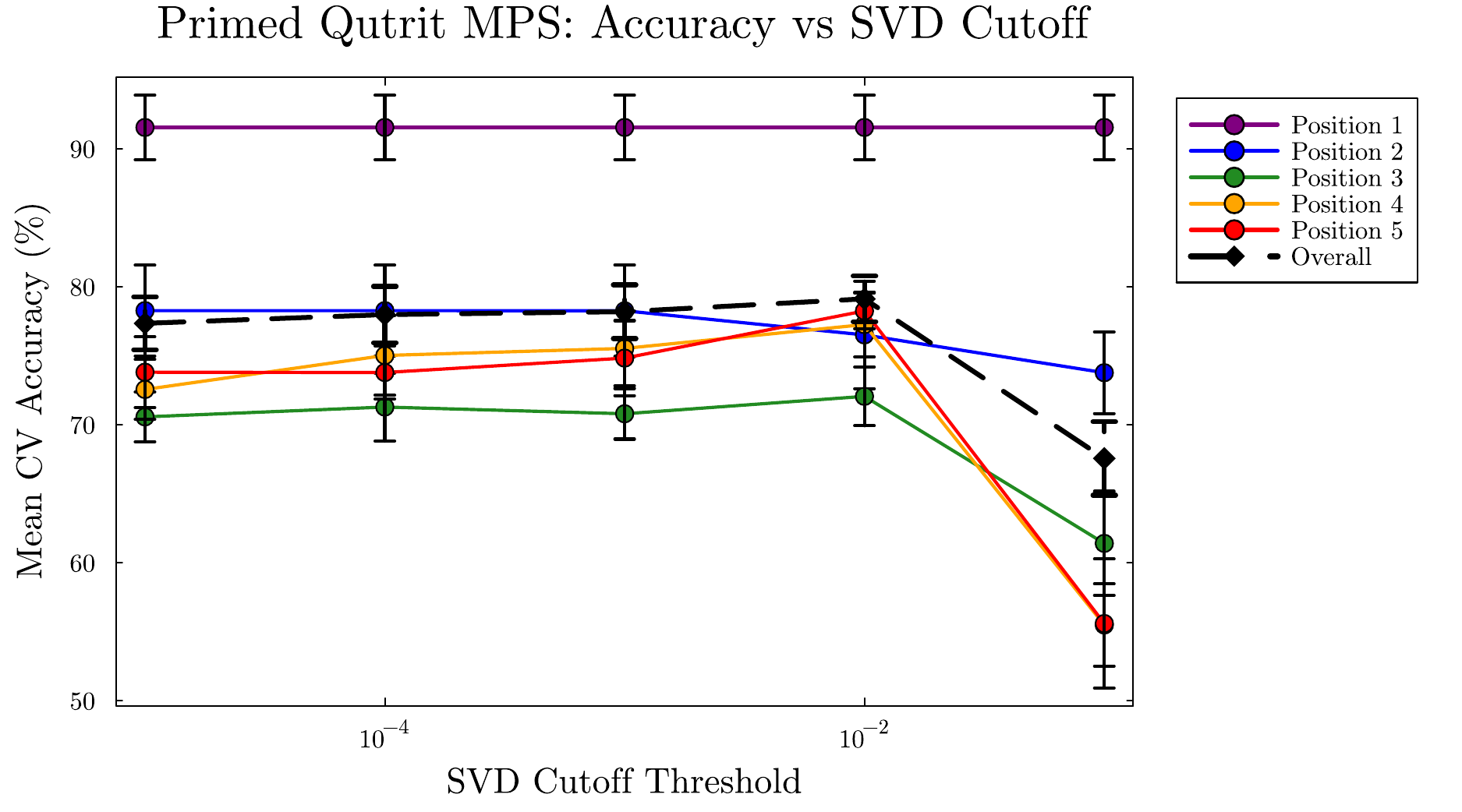}
	\end{subfigure}
	\end{center}
	\caption{Matrix product state cutoff $\epsilon$ sweep for qubit (\Cref{fig:qubitsweep}) and valence-primed qutrit (\Cref{fig:primedsweep}) models.}\label{fig:sweep}
\end{figure}

\textbf{Qubit model:} The empirical bond dimensions of [2, 4, 4, 2] confirms that correlation between recall events exist in the data and the model is able to encode them, instead of treating them as independent. A bond dimension of 1 at any cut would indicate that the recall outcomes on either side of that cut are statistically independent. The sharp decay profile in \Cref{fig:svdscree} validates that an MPS approximation is appropriate for this data, indicating that the probability structure is low-rank, and that the truncation introduced by SVD is discarding noise rather than structural signal. The symmetric pattern between bonds 1 and 4 and the larger dimensions at bonds 2 and 3 reflect the positional structure of the sequence, as the boundary items have fewer neighbours and therefore carry less correlation information through their virtual bonds. The higher bond dimensions and slower decay seen in the middle bonds is consistent with what would be expected from human memory, where the preceding outcomes affect the current item's recall. Additionally, the singular value in the second index increases as we progress through the bonds, implying that more and more information is being captured from previous events. 
However, the resultant bond dimensions are the maximum for a $d=2$ system at each cut, where the maximum possible bond dimension is $\text{min}(2^k, 2^{(n-k)})$. Every singular value above the noise floor is being retained, suggesting that the simplicity of the model and the sparsity of the data prevents meaningful compression of the network. The bond dimensions reported are as much a result of the data sparsity and threshold choice as they are of the underlying psychological behaviour. 

These results are indicative that the model has some capability to learn and understand the order dependencies present in human memory, and shows that with a larger and more complex model, more detail of the experimental data could be captured leading to improved prediction.

Varying the threshold value $\epsilon$ (\Cref{fig:qubitsweep}) shows that even extreme compression of the MPS results in little change to the model's accuracy, per position or overall. As the model has converged to predicting the majority class in the current position, there is no complex structure to lose under compression, even when bond dimensions reach [1,1,1,1].

\textbf{Qutrit model:} The singular value decay (\Cref{fig:svdscreequtrit}) is largely unchanged from the qubit model even with larger theoretical max dimensions. The empirical bond dimensions are [3, 5, 6, 3], higher than the qubit model (but below the theoretical maximum bond dimension of 9 for the central nodes). However, the actual decay does not support more information actually being encoded in the model. The asymmetry between bonds 1 and 4 and bonds 2 and 3 is more pronounced, showing that stronger connections are required between past and future as more information is accumulated across the trial.

The validity of using a tensor network approach is still apparent in this model, but the limitation appears to lie in the amount of data used to populate the tensor, rather than the structure of the model. However, by making a small structural change by introducing a valence priming stage before each recall site, we can drastically improve the predictive performance with the same amount and composition of data.

\textbf{Valence-primed qutrit model:} The empirical bond dimensions rise to [6, 22, 24, 6]. In all four bonds, the singular values decay much more gradually than previous models (\Cref{fig:svdscreeprimed}). The two central bonds in particular show a gradual decent rather than a sharp drop, with 8-10 components carrying significant weight. This indicates that the valence-conditioned joint distribution has a richer entanglement structure, with more singular components needed to represent the correlations between recall events at adjacent positions, and across the whole sequence. There is a pronounced asymmetry between bonds 1 and 4. Bond 4, which separates items 1–4 from item 5, shows a slower initial decay than bond 1, which is consistent with the idea that later items in the sequence are influenced by a richer accumulated context.

\textbf{Threshold Sweep:} Unlike previous models (see \Cref{fig:qubitsweep}), the valence-primed model is sensitive to the SVD cutoff threshold $\epsilon$ (\Cref{fig:primedsweep}). Performance is mostly stable until a threshold of $10^{-2}$, after which accuracy degrades, especially in the middle positions. This confirms that the higher-order singular values in the valence-primed MPS are carrying real predictive information, unlike the earlier models, where truncating those values had no effect due to contributing nothing to prediction. The middle positions (3 and 4) are affected the most (corresponding to bonds 2 and 3), further reinforcing that the many high index singular values of positions 2 and 3 are encoding entanglement information and are not just noise.

The sweep also suggests a practical point: thresholds between $1\times10^{-4}$ and $1\times10^{-2}$ give near-optimal accuracy while still compressing the network meaningfully, whereas increasing the threshold beyond $1\times10^{-2}$ degrades performance. This is the trade-off one would expect from an MPS applied to structured data~\cite{stoudenmire_supervised_2016}. The model has genuinely learnt the structures that are worth preserving.

The fact that neither the qubit nor qutrit models exhibit this behaviour, while the valence-primed model does, support the broader argument of this thesis: that emotional context is not peripheral to how memories are structured, but is woven into the dependencies between recall events in a way that a model must explicitly represent it to be effective.

\section{Discussion}\label{sec:Discussion}

In this Section, we will first summarise the work in this paper, before discussing limitations of both the study and the tensor network-based analysis presented here, then considering potential directions for future work.

\subsection{Summary of Findings}

This paper sets out three requirements for a model of emotional memory: representing the full distribution of recall probabilities across positions; sensitivity to the emotional valence of items and the order in which they appear; and the capacity to encode context-dependent structure between recall events.

The qubit and qutrit models each fulfil the first criterion, to varying degrees, with non-trivial bond dimensions providing evidence that recall outcomes are not statistically independent across positions.

The valence-primed model fulfils all three criteria. By encoding each memory event as a composite \textit{valence-recall} state, and conditioning on the known valence when predicting, we directly incorporate emotional valence into the probability structure which allows us to naturally capture how earlier valence-recall events shape future ones, without explicitly programming this (or any) behaviour into the model.
Incorporating emotional valence as a structural modelling variable rather than treating it as peripheral, or ignoring it completely, provides improvements in both the fidelity and predictive capability of tensor network models of memory.

Emotional valence and repetition shape children’s temporal order memory and can be interpreted within emerging quantum-like cognition frameworks. Consistent with predictions, positively-valenced objects were recalled more accurately than neutral objects in both object recall and temporal order accuracy. In contrast, no reliable RRE was observed, and recall performance remained stable across repeated trials. Overall, these findings suggest that emotional valence enhances the strength or accessibility of object and temporal representations, whereas repetition does not produce systematic improvements in recall performance, at least not over a short (approximately 5 minute) period of time. This pattern further supports accounts proposing that emotionally salient information receives enhanced attentional allocation during encoding~\cite{MatherSutherland2011}. It demonstrates robust valence effects, consistent with predictions, alongside minimal evidence of repetition-based change. Given the novelty in our approach and study design, these findings provide confidence that we are indeed accessing emotional memory processes. 

The findings collectively provide preliminary behavioural support for quantum-inspired models of cognition. Classical models assume stable memory traces, whereas quantum cognition proposes that cognitive states exist in probabilistic superpositions that collapse at retrieval~\cite{pothos_quantum_2022}. Within this framework, emotional information may alter the probability structure of recall. The advantage for positively-valenced objects may therefore reflect shifts in how event sequences are reconstructed, rather than encoding strength alone~\cite{BusemeyerWangTownsend2006}. A related concept is non-commutativity, whereby the order of information evaluation influences subsequent judgements~\cite{Wangetal2013}. In episodic recall, retrieving one element of a sequence may alter the probability of recalling subsequent elements. This suggests recall operates as a dynamic, state-dependent process, consistent with quantum principles, rather than retrieval of a fixed sequence, challenging classical accounts that emphasise strengthened encoding.

In the classical analysis (\Cref{sec:Class}), the observed enhancement of both object recall and temporal order memory for positively-valenced stimuli aligns with extensive evidence demonstrating that emotional information receives preferential processing and attention during encoding~\cite{MatherSutherland2011,TalmiPalombo2025,kensinger2009remembering}. Episodic memory requires binding of event components, including item identity and temporal relationships~\cite{KesnerHunsaker2010}. The simultaneous improvement in object recall and temporal order accuracy suggests that emotional salience strengthens the underlying episodic representation, rather than merely enhancing isolated item memory. Emotional salience may therefore also support the integration of ``what'' and ``when'' information during encoding, producing episodic representations that are more easily reconstructed during retrieval~\cite{Talmi2013}. In developmental terms, the enhanced recall of positively-valenced objects reflects an interaction between emotional salience and neural maturation. Temporal order memory is associated with coordinated activity between the hippocampus and PFC, regions that continue to develop throughout childhood~\cite{Bettencourtetal2021,Rigginsetal2016}. Children’s temporal memory is reconstructive rather than based on fixed sequential records~\cite{BrainerdReyna2005}. Emotional information may therefore serve as an effective organising cue during recall, guiding reconstruction of event sequences.

Preliminary patterns in the quantum-informed analyses indicate an asymmetry between neutral and positively-valenced items. Neutral items may place fewer cognitive demands, allowing subsequent items to be encoded with minimal interference. In contrast, positively-valenced items may require greater attentional resources. This pattern may reflect a cumulative effect of prior emotional context, whereby each successive item is evaluated within a dynamically changing cognitive state. From a quantum perspective, earlier items may progressively reshape the probability structure governing subsequent recall, consistent with interference effects predicted by quantum models. Within quantum cognition research, such effects relate to order-dependent phenomena, including QOE and RRE~\cite{Wangetal2013,OzawaKhrennikov2022}. Although no reliable RRE was observed, RRE represents only one form of order-dependent processing, and the observed asymmetry suggests broader interference effects through which emotional salience may shape temporal sequence reconstruction.  

These interpretations remain exploratory but suggest that temporal order memory reflects a emotionally-sensitive process. Rather than replacing traditional accounts, quantum-like cognition provides a complementary framework for modelling how emotional and contextual factors dynamically shape recall. Ongoing work with computer science collaborators is developing formal quantum models to test these mechanisms formally.

Research applying quantum cognition to developmental memory remains limited, with most work focusing on adult decision-making~\cite{pothos_quantum_2022}. The present study therefore contributes to an emerging interdisciplinary area, demonstrating how children’s recall may be interpreted within alternative formal frameworks while remaining grounded in established developmental theory. 

\subsection{Limitations}
We here discuss the limitations of the experiment, the data generated, and the tensor networks used to model this data.

\textbf{Data Sparsity}:
Whilst not a significant problem for the qubit model, for the qutrit and valence-primed models there is a large mismatch between dataset size and their respective state spaces, with both reliant on the Laplace smoothing to fill in missing patterns of recall. An example of this is that only 4\% of the recall outcomes across all trials and positions were completely forgotten (as opposed to remembered out of order), meaning at least the qutrit model fails to capture any meaningful information from this limited data. In this specific study, participants were able to recall the items they saw the majority of the time, in-order or not. The strong performance of the valence-primed model seems to imply this is not a significant issue for it, but it should be interpreted with the caveat that a larger dataset with more varied recall patterns would provide a more robust test of the model, and of Tensor Network based models in general.

\textbf{Single dataset and emotional stimuli}:
All three models were evaluated on a single dataset from one type of experiment. The generalisability of the findings to other memory tasks, stimulus types, or participant populations is therefore unknown. A database of emotionally-valenced, tactile toys does not exist. These data suggest that the stimuli used did evoke emotional valence but the limitations of this approach must be acknowledged. 

\textbf{Prediction Conditioning}:
The prediction procedure conditions on past recall outcomes to predict the current item, which assumes the model has access to a participant's recall history up to that point. In a real clinical or applied setting this information may not always be available in the required form, and the sequential nature of the conditioning means that early prediction errors could propagate and degrade performance on later items. This was not investigated here and represents a practical limitation on the deployability of the approach.

\textbf{State Space Scaling}:
Despite modelling a simple memory task with 5 items, 3 recall options and 2 valences, the size of the state space required to represent every combination of outcome is still 7,776. This number scales exponentially with the number of items, and polynomially with the number of recall outcomes. As discussed above, Matrix Product States resolve this exponential growth by truncating the global probability tensor. However, there is no guarantee that an MPS approach will remain appropriate as chain length or physical index size grows, and that SSVD will continue to be able to compress the tensor effectively. As sequence length increases, the entanglement structure between distant items may become too rich and complex for a low-rank MPS approximation to capture, requiring bond dimensions that grow prohibitively large. This represents a fundamental scalability question that would need to be addressed before tensor network models of this kind could be applied to more realistic memory tasks.

\subsection{Future Directions}

Contrary to the prediction that performance would change across trials, recall accuracy remained relatively stable. The absence of a trial effect suggests task accessibility rather than an absence of learning, indicating participants acquired task demands rapidly in the first trials. This interpretation closely aligns with developmental findings showing stable performance when encoding conditions are sufficient for integration~\cite{Gathercoleetal2004}. Future work must balance task demands, and here we provide evidence of a task that effectively assess emotional memory in children. 

The classification of stimuli as positively-valenced relied on researcher judgement, despite emotional responses to stimuli being inherently subjective~\cite{LangBradley2010}. Informal observations indicated variability in children’s engagement with stimuli, highlighting the challenge of operationalising emotional valence in developmental populations. Future studies could incorporate independent valence ratings or validated child-appropriate stimulus sets~\cite{Talmi2013}. Abstract sequencing tasks may also help isolate temporal reconstruction processes from object preferences.

The physical nature of the task required children to manipulate objects when reconstructing sequences. Although this introduces variability related to motor coordination and attention, it enhances applicability to real-life settings (ecological validity) by approximating real-world memory reconstruction. This approach is particularly appropriate in developmental research, where physically interactive tasks can improve comprehension and sustain attention~\cite{Diamond2013}. Ecological validity in tasks is an important consideration when developing quantum-like models of cognition, it ensures that the most value and impact can be gained by translating tools generated to other cognitive tasks and processes. 

Future work could explore non-cognitive measures, such as physiological responses (e.g. galvanic skin conductance; eye movements and pupil dilation) collected in tandem with cognitive judgements. Quantum-like models of emotional stimuli have been proposed~\cite{Whiteetal2016}, and replication of the models we give in the present paper with the addition of physiological indices and a repeated approach to measure RRE would allow a point of comparison to explore these models further.

As discussed above, emotional information may serve as an effective organising cue during recall, guiding reconstruction of event sequences. Future research could therefore examine the enhanced recall for positive items using validated emotional stimulus sets and independent valence ratings from children to determine whether the observed enhancement reflects perceived emotional engagement rather than researcher-defined stimulus categories.

In \Cref{sec:Class}, we briefly mention that we do not observe RRE. However, this is in truth a simplification - unlike previous quantum cognition analysis, which has relied on ``survey questions'' designed to judge attitudes towards something (like the Gore-Clinton poll), in the current case there is a ``correct'' answer. Therefore, it is not trivial to say whether RRE would show up as an improvement in scores (due to repeated priming), or as an expectation that, on a second presentation of a given set, participants would make mistakes \emph{in exactly the same way} as they did the first time around. Future works could try to disambiguate these, and come up with more nuanced versions of RRE applicable to such a ``testing'' situation.

\subsection{Conclusion}
In summary, emotional salience enhances children’s object recall and temporal memory, whereas repetition does not produce systematic changes in performance. These findings indicate that children’s temporal memories are reconstructed in a context-sensitive manner rather than retrieved as fixed sequences, highlighting the dynamic nature of recognition and temporal memory development. By integrating behavioural findings with emerging theoretical accounts, this study provides preliminary evidence that quantum cognition offers a coherent framework for modelling how probabilistic cognitive states may give rise to specific recall outcomes during retrieval. More broadly, these findings suggest that understanding children’s memory requires models adequately that capture context sensitivity, interference, and state-dependent processing, positioning quantum-inspired approaches as a promising direction for future developmental research. 

\textit{Acknowledgements -} We thank the families who participated in the study Super Sequencing at the Science Adventures outreach days at Newcastle University, and also Zak Mala who assisted with data collection. B-AR and JRH acknowledge support from a UKRI Cross Research Council Responsive Mode Grant (UKRI3740). JRH acknowledges support from a Royal Society Research Grant (RG/R1/251590), and an EPSRC Mathematical Sciences Small Grant (UKRI3647). HG and JRH acknowledge support from JRH's EPSRC Quantum Technologies Career Acceleration Fellowship (UKRI1217). 

\bibliographystyle{unsrturl}
\bibliography{ref.bib}

\appendix
\section{Area Law Violation}

MPSs were originally developed in the context of the Density Matrix Renormalisation Group (DMRG) algorithm in quantum physics, which is used to find the ground state of 1-dimensional quantum systems~\cite{schollwock_density-matrix_2011}. These real-life systems obey the \textit{entanglement area law}, which states that the entanglement entropy in a system is bounded by a constant, rather than growing with the size of the system as one would expect. This law guarantees that an MPS representing this physical state, can represent the state efficiently and accurately with a small, fixed bond dimension~\cite{eisert_colloquium_2010, hastings_area_2007}.
It is important to note that human memory data most likely violates this law, and therefore this guarantee does not necessarily carry over to the use of tensor networks to model human memory. The tensor constructed from the experimental data is not a quantum ground state, it has no underlying Hamiltonian, with no energy minimisation. Therefore, there is no reason to expect the area law to hold. Without this guarantee, we had to empirically determine whether an MPS approximation is valid, by analysing each step when constructing the MPS using scree plots, ensuring that the process only introduces small errors.

Thankfully, the validity of the MPS approximation was justified empirically here through the singular value scree plots, which show rapid decay in all three models. However, the scree plots only tell us that the global structure of the tensor is low-rank, and say nothing about whether the approximation is equally faithful across all entries. In a sparse tensor like the valence-primed model's, where many sequences are rarely or never observed, the SVD truncation could be introducing larger errors in the areas where the data is sparse, which would not be visible in the decay profile.

\end{document}